\documentclass[smallextended]{svjour3}       
\smartqed  
\usepackage{graphicx}

\usepackage{graphics, subfigure, theorem, times, amsfonts, amsmath, amssymb, cite, verbatim, math tools}
\usepackage{tikz}
\usepackage{natbib}
\usepackage{framed}
\usetikzlibrary{shapes,arrows}
\tikzstyle{phantom vertex} = [ ellipse, 
                               anchor = center, 
                               minimum height = 1*\unit, 
                               minimum width  = 1*\unit,
                               inner sep=0pt,
                               anchor=center]
\tikzstyle{red vertex}   = [black, fill = red!20,   phantom vertex, draw]
\tikzstyle{black vertex} = [black, fill = black!20, phantom vertex, draw]
\tikzstyle{blue vertex}  = [black, fill = blue!20,  phantom vertex, draw]
\tikzstyle{green vertex} = [black, fill = green!20,  phantom vertex, draw]
\tikzstyle{vertex}       = [draw, phantom vertex]

\tikzstyle{point} = [ellipse, inner sep=0pt, draw, fill=white, anchor = center,
                     minimum height = 0.05*\unit, minimum width  = 0.05*\unit]

\usepackage{pgfplots}
\usepackage{color}
\usepackage{needspace}

\newcommand{\myindentedparagraph}[1]{\needspace{1\baselineskip}\medskip \hangindent=11pt \hangafter=0 \noindent{\it #1.}}

\usepackage{hyperref}
\usepackage{flushend}
\usepackage{multirow}
\usepackage{rotating}
\usepackage{framed}
\input{mysymbol.sty}

\def \mlc {\text{\normalfont mlc}}
\def \sep {\text{\normalfont sep}}

\def\R {\text{\normalfont R}}

\def\NR {\text{\normalfont NR}}

\def\SL {\text{\normalfont SL}}
\def\U {\text{\normalfont U}}

\def\N{\mathbb{N}}

\newcommand{\QED}{\hfill\ensuremath{\blacksquare}}
\newtheorem{myclaim}{\hspace{0pt}\bf Claim}

\AtBeginDocument{%
  \paperwidth=\dimexpr
    1in + \oddsidemargin
    + \textwidth
    + 1in + \oddsidemargin
  \relax
  \paperheight=\dimexpr
    1in + \topmargin
    + \headheight + \headsep
    + \textheight
    + 1in + \topmargin
  \relax
  \usepackage[pass]{geometry}\relax
}

\begin{document}

\title{Hierarchical Clustering of Asymmetric Networks \thanks{Work in this paper is supported by NSF CCF-1217963, NSF CAREER CCF-0952867, NSF IIS-1422400, NSF CCF-1526513, AFOSR FA9550-09-0-1-0531, AFOSR FA9550-09-1-0643, NSF DMS-0905823, and NSF DMS-0406992.}
}


\author{Gunnar Carlsson   \and
	Facundo M\'emoli 	\and
	Alejandro Ribeiro 	\and
	Santiago Segarra
}

\authorrunning{ } 

\institute{G. Carlsson \at
              Department of Mathematics, Stanford University \\
              \email{gunnar@math.stanford.edu}            \and
              F. M\'emoli  \at
              Department of Mathematics and Department of Computer Science and Engineering, Ohio State University \\                            	      \email{memoli@math.osu.edu}            \and
              A. Ribeiro and S. Segarra \at
	Department of Electrical and Systems Engineering, University of Pennsylvania \\
	\email{aribeiro@seas.upenn.edu, ssegarra@seas.upenn.edu} 
}

\date{Received: date / Accepted: date}

\maketitle

\begin{abstract}
This paper considers networks where relationships between nodes are represented by directed dissimilarities. The goal is to study methods that, based on the dissimilarity structure, output hierarchical clusters, i.e., a family of nested partitions indexed by a connectivity parameter. Our construction of hierarchical clustering methods is built around the concept of admissible methods, which are those that abide by the axioms of value -- nodes in a network with two nodes are clustered together at the maximum of the two dissimilarities between them -- and transformation -- when dissimilarities are reduced, the network may become more clustered but not less. Two particular methods, termed reciprocal and nonreciprocal clustering, are shown to provide upper and lower bounds in the space of admissible methods. Furthermore, alternative clustering methodologies and axioms are considered. In particular, modifying the axiom of value such that clustering in two-node networks occurs at the minimum of the two dissimilarities entails the existence of a unique admissible clustering method.
\end{abstract}

\section{Introduction}\label{sec_introduction}

The problem of determining clusters in a data set admits different interpretations depending on whether the underlying data is metric, symmetric but not necessarily metric, or asymmetric. Of these three classes of problems, clustering of metric data is the most studied one in terms of both, practice and theoretical foundations. In terms of practice there are literally hundreds of methods, techniques, and heuristics that can be applied to the determination of hierarchical and nonhierarchical clusters in finite metric spaces -- see, e.g., \citet{RuiWunsch05}. Theoretical foundations of clustering methods, while not as well developed as their practical applications \citep{vonlux-david, sober,science_art}, have been evolving over the past decade \citep{ben-david-ackermann,ben-david-reza, CarlssonMemoli10, kleinberg, clust-um,multi-param}. Of particular relevance to our work is the case of hierarchical clustering where, instead of a single partition, we look for a family of partitions indexed by a resolution parameter; see e.g., \citet{lance67general, clusteringref}. In this context, it has been shown by \citet{clust-um} that single linkage \citep[Ch. 4]{clusteringref} is the unique hierarchical clustering method that satisfies three reasonable axioms. These axioms require that the hierarchical clustering of a metric space with two points is the same metric space, that there be no non-singleton clusters at resolutions smaller than the smallest distance in the space, and that when distances shrink, the metric space may become more clustered but not less.

When we remove the condition that the data be metric, we move into the realm of clustering in weighted networks, i.e. a set of nodes with pairwise and possibly \emph{directed} dissimilarities represented by edge weights. For the undirected case, the knowledge of theoretical underpinnings is incipient but practice is well developed. Determining clusters in this undirected context is often termed community detection and is formulated in terms of finding cuts such that the edges between different groups have high dissimilarities -- meaning points in different groups are dissimilar from each other -- and the edges within a group have small dissimilarities -- which means that points within the same cluster are similar to each other \citep{ShiMalik00, GirvanNewman02, GirvanNewman04}. An alternative approach for clustering nodes in graphs is the idea of spectral clustering \citep{Chung97, spectral-clustering, NgEtal02, BachJordan03}. When a graph contains several connected components its Laplacian matrix has multiple eigenvectors associated with the null eigenvalue and the nonzero elements of the corresponding eigenvectors identify the different connected components. The underlying idea of spectral clustering is that different communities should be identified by examining the eigenvectors associated with eigenvalues close to zero. 

Further relaxing symmetry so that we can allow for asymmetric relationships between nodes \citep{SaitoYadohisa04} reduces the number of available methods that can deal with such data \citep{hubert-min,slater1976hierarchical,boyd-asymmetric,tarjan-improved,slater1984partial,murtagh-multidimensional,PentneyMeila05, MeilaPentney07, ZhouEtal05}. Examples of these methods are the adaptation of spectral clustering to asymmetric graphs by using a random walk perspective \citep{PentneyMeila05} and the use of weighted cuts of minimum aggregate cost \citep{MeilaPentney07}. In spite of these contributions, the rarity of clustering methods for asymmetric networks is expected because the interpretation of clusters as groups of nodes that are closer to each other than to the rest is difficult to generalize when nodes are close in one direction but far apart in the other. 

Although it is difficult to articulate a general intuition for clustering of asymmetric networks, there are nevertheless some behaviors that we should demand from any reasonable clustering method. Following \citet{kleinberg} and \citet{clust-um}, the perspective taken in this paper is to impose these desired behaviors as axioms and proceed to characterize the space of methods that are admissible with respect to them. While different axiomatic constructions are discussed here, the general message is that strong structure can be induced by seemingly weak axioms.

In Section~\ref{sec_preliminaries} we introduce notions related to network theory and clustering needed for the development of the results presented in this paper. In particular, we revisit the known equivalence between dendrograms and ultrametrics (Section~\ref{sec_dendrograms_and_ultrametrics}), which is instrumental to our proofs. The axioms of value and transformation are stated formally in Section~\ref{sec_axioms} but they correspond to the following intuitions:

\myindentedparagraph{(A1) Axiom of Value} For a network with two nodes, the nodes are clustered together at a resolution equal to the maximum of the two intervening dissimilarities. 

\myindentedparagraph{(A2) Axiom of Transformation} If we consider a domain network and map it into a target network in a manner such that no pairwise dissimilarity is increased by the mapping, then the resolution level at which two nodes in the target network become part of the same cluster is not larger than the level at which they were clustered together in the original domain network.

\vspace{0.075in}\noindent A hierarchical clustering method satisfying axioms (A1) and (A2) is said to be {\it admissible.}
Our first theoretical study is the relationship between clustering and mutual influence in networks of arbitrary size (Section \ref{sec_axiomatic_structure}). In particular, we show that the outcome of any admissible hierarchical clustering method is such that a necessary condition for two nodes to cluster together is the existence of chains that allow for direct or indirect influence between the nodes. Two hierarchical clustering methods that abide by axioms (A1) and (A2) are derived in Section~\ref{sec_reicprocal_and_nonreciprocal}. The first method, \emph{reciprocal clustering}, requires clusters to form through edges exhibiting low dissimilarity in both directions whereas the second method, \emph{nonreciprocal clustering}, allows clusters to form through cycles of small dissimilarity. 
A fundamental result regarding admissible methods is the proof that any clustering method that satisfies axioms (A1) and (A2) lies between reciprocal and nonreciprocal clustering in a well-defined sense (Section \ref{sec_extremal_ultrametrics}). Specifically, any clustering method that satisfies axioms (A1) and (A2) forms clusters at resolutions larger than the resolutions at which they are formed with nonreciprocal clustering, and smaller than the resolutions at which they are formed with reciprocal clustering. When restricted to symmetric networks, reciprocal and nonreciprocal clustering yield equivalent outputs, which coincide with the output of single linkage (Section \ref{secsymmetric_networks}). This observation is consistent with the existence and uniqueness result by \citet{clust-um} since axioms (A1) and (A2) are reduced to two of the axioms considered there when we restrict attention to metric data. The derivations in our paper show that the existence and uniqueness result by \citet{clust-um} is true for all symmetric, not necessarily metric, datasets and that a third axiom considered there is redundant because it is implied by the other two.

In some applications the requirement for bidirectional influence in the Axiom of Value is not justified as unidirectional influence suffices to establish proximity. This alternative value statement leads to the study of alternative axiomatic constructions and their corresponding admissible hierarchical clustering methods (Section \ref{sec_alternative_axioms}). We first propose an Alternative Axiom of Value in which clusters in two-node networks are formed at the minimum of the two dissimilarities. Under this axiomatic framework we define unilateral clustering as a method in which influence propagates through chains of nodes that are close in at least one direction (Section \ref{sec_unilateral_clustering}). Contrary to the case of admissibility with respect to (A1)-(A2) in which a range of methods exist, unilateral clustering is the unique method that is admissible with respect to the Alternative Axiom of Value. Lastly, an agnostic position where nodes in two-node networks are allowed to cluster at any resolution between the minimum and the maximum dissimilarity between them is also studied (Section \ref{sec_agnostic_axiom_of_value}).

Conclusive remarks are presented in Section~\ref{sec_conclusions}. All proofs not included in the main body of the text can be found in the Appendix (Section~\ref{sec_appendix}).

\section {Preliminaries}\label{sec_preliminaries}

We define a network $N_X$ to be a pair $(X, A_X)$ where $X$ is a finite set of points or nodes and $A_X: X \times X \to \reals_+$ is a dissimilarity function. The dissimilarity $A_X(x,x')$ between nodes $x\in X$ and $x'\in X$ is assumed to be non-negative for all pairs $(x,x')$ and 0 if and only if $x=x'$. We do not, however, require $A_X$ to be a metric on the finite set $X$. Specifically, dissimilarity functions $A_X$ need not satisfy the triangle inequality and, more consequential for the problem considered here, they may be asymmetric in that it is possible to have $A_X(x,x')\neq A_X(x',x)$ for some $x \neq x'$. We further define $\ccalN$ as the set of all networks $N_X$. Networks in $\ccalN$ can have different node sets $X$ as well as different dissimilarity functions $A_X$.

The smallest non-trivial networks contain two nodes $p$ and $q$ and two dissimilarities $\alpha$ and $\beta$ as depicted in Fig. \ref{fig_axioms_value_influence}. The following special networks appear often throughout our paper: consider the dissimilarity function $A_{p,q}$ with $A_{p,q}(p,q)=\alpha$ and $A_{p,q}(q,p)=\beta$ for some $\alpha, \beta >0$ and define the {\it two-node network} $\vec{\Delta}_2(\alpha, \beta)$ with parameters $\alpha$ and $\beta$ as $\vec{\Delta}_2(\alpha, \beta):= (\{p,q\}, A_{p,q})$. 

By a clustering of the set $X$ we mean a partition $P_X$ of $X$; i.e., a collection of sets $P_X=\{B_1,\ldots, B_J\}$ which are pairwise disjoint, $B_i\cap B_j =\emptyset$ for $i\neq j$, and are required to cover $X$, $\cup_{i=1}^{J} B_i = X$. The sets $B_1, B_2, \ldots B_J$ are called the \emph{blocks} or \emph{clusters} of $P_X$. We define the \emph{power set} $\ccalP(X)$ of $X$ as the set containing every subset of $X$, thus $B_i \in \ccalP(X)$ for all $i$. An equivalence relation $\sim $ on $X$ is a binary relation such that for all $x, x', x'' \in X$ we have that (1) $x \sim x$, (2) $x \sim x'$ if and only if $x' \sim x$, and (3)  $x \sim x'$  and $x' \sim x''$ imply $x \sim x''$.

A partition $P_X=\{B_1,\ldots, B_J\}$ of $X$ induces and is induced by an equivalence relation $ \sim_{P_X} $ on $X$ where, for all $x, x' \in X$, we have that $x \sim_{P_X} x'$ if and only if $x$ and $x'$ belong to the same block. In this paper we focus on hierarchical clustering methods. The output of hierarchical clustering methods is not a single partition $P_X$ but a nested collection $D_X$ of partitions $D_X(\delta)$ indexed by a resolution parameter $\delta\geq 0$. In consistency with our previous notation, for a given $D_X$, we say that two nodes $x$ and $x'$ are equivalent at resolution $\delta \geq 0$ and write $x\sim_{D_X(\delta)} x'$ if and only if nodes $x$ and $x'$ are in the same block of $D_X(\delta)$. The nested collection $D_X$ is termed a \emph{dendrogram} and is required to satisfy the following two properties plus a technical condition \citep{clust-um}: 

\myindentedparagraph{(D1) Boundary conditions} For $\delta=0$ the partition $D_X(0)$ clusters each $x\in X$ into a separate singleton and for some $\delta_0$ sufficiently large $D_X(\delta_0)$ clusters all elements of $X$ into a single set,
%
 $  D_X(0)  = \big\{ \{x\}, \, x\in X\big\},  \quad D_X(\delta_0) = \big\{ X \big\} \quad \forsome\ \delta_0 > 0$.

\myindentedparagraph{(D2) Hierarchy} As $\delta$ increases clusters can be combined but not separated. I.e., for any $\delta_1 < \delta_2$ any pair of points $x,x'$ for which $x\sim_{D_X(\delta_1)} x'$ must be  $x\sim_{D_X(\delta_2)} x'$.

\vspace{0.075in}\noindent The interpretation of a dendrogram is that of a structure which yields different clusterings at different resolutions. At resolution $\delta=0$ each point is in a cluster of its own. As the resolution parameter $\delta$ increases, nodes start forming clusters. According to condition (D2), nodes become ever more clustered since once they join together in a cluster, they stay together in the same cluster for all larger resolutions. Eventually, the resolutions become coarse enough so that all nodes become members of the same cluster and stay that way as $\delta$ keeps increasing. A dendrogram can be represented as a rooted tree; see e.g. Fig.~\ref{fig_dendrograms_as_ultrametrics}.

Denoting by $\ccalD$ the space of all dendrograms we define a hierarchical clustering method as a function 
\begin{equation}\label{eqn_clust_from_networks_to_dendrograms}
\ccalH:\ccalN \to \ccalD,
\end{equation}
from the space of networks $\ccalN$ to the space of dendrograms $\ccalD$ such that the underlying node set $X$ is preserved. For the network $N_X=(X,A_X)$ we denote by $D_X=\ccalH(X,A_X)$ the output of clustering method $\ccalH$. 

In the description of hierarchical clustering methods the concepts of \emph{chain}, \emph{chain cost}, and \emph{minimum chain cost} are important. Given a network $(X, A_X)$ and $x, x' \in X$, a chain from $x$ to $x'$ is any \emph{ordered} sequence of nodes $[x=x_0, x_1, \ldots , x_{l-1}, x_l=x']$, which starts at $x$ and finishes at $x'$. We will frequently use the notation $C(x,x')$ to denote one such chain. We say that $C(x, x')$ links or connects $x$ to $x'$. Given two chains $C(x, x')=[x=x_0, x_1, ... , x_l=x']$ and $C(x', x'')=[x'=x'_0, x'_1, ... , x'_{l'}=x'']$ such that the end point of the first one coincides with the starting point of the second one, we define the \emph{concatenated chain} $C(x, x') \uplus C(x',x'')$ as
\begin{align}\label{eqn_definition_concatenation}
   C(x, x') \uplus C(x',x'') := [x=x_0,\ldots , x_l=x'=x'_0,\ldots , x'_{l'}=x'']. 
\end{align}
Observe that the chain $C(x, x')=[x=x_0, x_1, \ldots , x_{l-1}, x_l=x']$ and its reverse $[x'=x_l, x_{l-1}, \ldots , x_{1}, x_0=x]$ are different entities even if the intermediate hops are the same. The \emph{links} of a chain are the edges connecting its consecutive nodes in the direction imposed by the chain. We define the \emph{cost} of a given chain $C(x, x')=[x=x_0,\ldots, x_l=x']$ as $\max_{i | x_i\in C(x,x')}A_X(x_i,x_{i+1})$, i.e., the maximum dissimilarity encountered when traversing its links in order. The directed minimum chain cost $\tdu^*_X(x, x')$ between $x$ and $x'$ is then defined as the minimum cost among all the chains connecting $x$ to $x'$,
\begin{align}\label{eqn_nonreciprocal_chains} 
   \tdu^*_X(x, x') := \min_{C(x,x')} \,\,
                        \max_{i | x_i\in C(x,x')} A_X(x_i,x_{i+1}).
\end{align} 
In asymmetric networks the minimum chain costs $\tdu^*_X(x, x')$ and $\tdu^*_X(x', x)$ are different in general but they are equal on symmetric networks. In this latter case, the costs $\tdu^*_X(x, x') = \tdu^*_X(x', x)$ are instrumental in the definition of single linkage clustering \citep{clust-um}. Indeed, for resolution $\delta$, single linkage makes $x$ and $x'$ part of the same cluster if and only if they can be linked through a chain of cost not exceeding $\delta$. Formally, the equivalence classes at resolution $\delta$ in the single linkage dendrogram $\text{SL}_X$ over a symmetric network $(X, A_X)$ are defined by
\begin{equation}\label{eqn_single_linkage}
   x\sim_{\text{SL}_X(\delta)} x' \iff  
      \tdu^*_X(x, x') = \tdu^*_X(x', x) \leq\delta.
\end{equation}

We further define a \emph{loop} as a chain of the form $C(x,x)$ for some $x \in X$ such that $C(x, x)$ contains at least one node other than $x$. Since a loop is a particular case of a chain, the cost of a loop is given by its largest dissimilarity. Furthermore, consistently with \eqref{eqn_nonreciprocal_chains}, we define the \emph{minimum loop cost} $\mlc(X,A_X)$ of a network $(X, A_X)$ as the minimum across all possible loops of each individual loop cost,
\begin{equation}\label{eqn_def_mlc}
    \mlc(X,A_X):=\min_x \, \min_{C(x,x)} \,\,  \max_{i | x_i\in C(x,x)}A_X(x_i,x_{i+1}),
\end{equation}
where, we recall, $C(x,x)$ contains at least one node different from $x$. Another relevant property of a network $(X, A_X)$ is the \emph{separation} of the network $\sep(X,A_X)$ which we define as its minimum positive dissimilarity, 
\begin{equation}\label{eqn_def_separation_network}
   \sep(X,A_X) := \min_{x \neq x'} A_X(x, x').
\end{equation}
Notice that from \eqref{eqn_def_mlc} and \eqref{eqn_def_separation_network} we must have $\sep(X,A_X) \leq \mlc(X,A_X)$. Further observe that in the particular case of networks with symmetric dissimilarities the two quantities coincide, i.e., $\sep(X,A_X)=\mlc(X,A_X)$.

When one restricts attention to networks $(X,A_X)$ having dissimilarities $A_X$ that conform to the definition of a finite metric space -- i.e., dissimilarities $A_X$ are symmetric and satisfy the triangle inequality -- it has been shown by \citet{clust-um} that single linkage is the unique hierarchical clustering method satisfying axioms (A1)-(A2) in Section \ref{sec_axioms} plus a third axiom stating that clusters cannot form at resolutions smaller than the minimum distance between different points of the space. In the case of asymmetric networks the space of admissible methods is richer, as we demonstrate throughout this paper.

\subsection {Dendrograms as ultrametrics}\label{sec_dendrograms_and_ultrametrics}

Dendrograms are convenient graphical representations but otherwise cumbersome to handle. A mathematically more convenient representation is obtained when one identifies dendrograms with finite \emph{ultrametric} spaces. An ultrametric defined on the set $X$ is a metric function $u_X: X \times X \to \reals_+$ that satisfies a stronger triangle inequality as we formally define next.

%
\begin{definition}\label{def_ultrametric} Given a node set $X$, an ultrametric $u_X$ is a non-negative function $u_X: X \times X \to \reals_+$ satisfying the following properties:
\begin{mylist}
\item[{\it (i) Identity.}] The ultrametric $u_X(x, x')=0$ if and only if $x=x'$ for all $x, x' \in X$.
\item[{\it (ii) Symmetry.}] For all pairs of points $x,x'\in X$ it holds that $u_X(x, x') \! = \! u_X(x', x)$.
\item[{\it (iii) Strong triangle inequality.}] Given $x,x',x''\in X$, the ultrametrics $u_X(x,x'')$, $u_X(x,x')$, and $u_X(x',x'')$ satisfy the strong triangle inequality
\begin{equation}\label{eqn_strong_triangle_inequality}
    u_X(x,x'') \leq \max \Big(u_X(x,x'),  u_X(x',x'') \Big).
\end{equation} \end{mylist}\vspace{-10pt} \end{definition}

\noindent Since \eqref{eqn_strong_triangle_inequality} implies the usual triangle inequality $u_X(x,x'') \leq u_X(x,x') + u_X(x',x'')$ for all $x, x', x'' \in X$, ultrametric spaces are particular cases of metric spaces.

\begin{figure}
\centering
\def \thisplotscale {0.6}
\def \unit {\thisplotscale cm}

\usetikzlibrary{positioning}

{\small
\begin{tikzpicture}[thick, x = 1.4*\unit, y = 1.0*\unit]

    \coordinate (1 end) at (2.0, 1.0); 
    \coordinate (2 end) at (2.0, 2.0); 
    \coordinate (3 end) at (4.0, 3.0); 
    \coordinate (4 end) at (4.0, 4.0); 
    \coordinate (1 and 2 end) at (6.0, 1.5); 
    \coordinate (3 and 4 end) at (6.0, 3.5); 
    \coordinate (1 2 3 and 4 end) at (7.0, 2.5); 

    \coordinate (y-axis-begin) at (0, 0.5); 
    \coordinate (y-axis-end) at (0, 5.2); 
    \coordinate (x-axis-begin) at (0, 0.5); 
    \coordinate (x-axis-end) at (8, 0.5); 

    \path[draw, thin, -stealth] (y-axis-begin) -- (y-axis-end); 
    \path[draw, thin, -stealth] (x-axis-begin) -- (x-axis-end); 

    \path[draw, thin, dashed] (x-axis-begin) ++ (1,0) node [below] {{$1$}}
    ++ (1,0) node [below] {{$2$}}
    ++ (1,0) node [below] {{$3$}}
    ++ (1,0) node [below] {{$4$}}
    ++ (1,0) node [below] {{$5$}}
    ++ (1,0) node [below] {{$6$}}
    ++ (1.5,-0.1) node [below] {{Resolution $\delta$}}    
    ;
    
    \coordinate (first) at (-2.2, 2.5); 
	\node [right=of first,rotate=90,anchor=north] (first) {Nodes};

    \path[draw, thick] (-0., 1) node[left] {$a$} -- (1 end); 
    \path[draw, thick] (-0., 2) node[left] {$b$} -- (2 end); 
    \path[draw, thick] (-0., 3) node[left] {$c$} -- (3 end); 
    \path[draw, thick] (-0., 4) node[left] {$d$} -- (4 end); 

    \path[draw, thick] (1 end) -- (2 end); 
    \path[draw, thick] (3 end) -- (4 end); 

    \path[draw, thick] (1 end) ++ (0,0.5) -- (1 and 2 end); 
    \path[draw, thick] (3 end) ++ (0,0.5) -- (3 and 4 end); 

    \path[draw, thick] (1 and 2 end) -- (3 and 4 end); 

    \path[draw, thick] (1 and 2 end) ++ (0,1) -- (1 2 3 and 4 end); 

    \path[draw, thin, dashed] (2 end) -- ++ (0,2.5) 
                                ++ (-0.4,0) node [above] {{$u_X(a,b)=2$}};
    \path[draw, thin, dashed] (4 end) -- ++ (0,0.5) 
                                ++ (-0.,0) node [above] {{$u_X(c,d)=4$}};
    \path[draw, thin, dashed] (3 and 4 end) -- ++ (0,1.0) 
                                ++ (0.4,0) node [above] {{$u_X(a,c)=6$}};
           
\end{tikzpicture}
}
\vspace{-0.05in}
\caption{Equivalence of dendrograms and ultrametrics. Given a dendrogram $D_X$ define the function $u_X(x,x') := \min \big\{ \delta \geq 0 \, | \, x\sim_{D_X(\delta)} x' \big\}$. This function is an ultrametric because it satisfies the identity property, the strong triangle inequality \eqref{eqn_strong_triangle_inequality} and is symmetric.}
\vspace{-0.1in}
\label{fig_dendrograms_as_ultrametrics}
\end{figure}
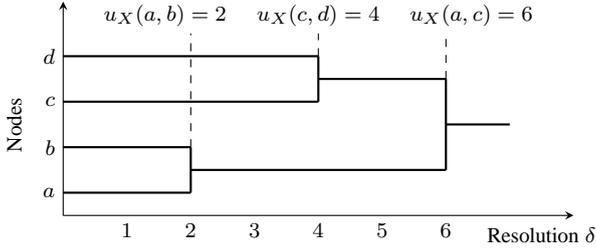

Our interest in ultrametrics stems from the fact that it is possible to establish a structure preserving bijective mapping between dendrograms and ultrametrics as proved by the following construction and theorem; see also Fig. \ref{fig_dendrograms_as_ultrametrics}.

%

Consider the map $\Psi:\mathcal{D} \rightarrow \mathcal{U}$ from the space of dendrograms to the space of networks endowed with ultrametrics, defined as follows: for a given dendrogram $D_X$ over the finite set $X$ write $\Psi(D_X) = (X, u_X)$, where we define $u_X(x,x')$ for all $x, x' \in X$ as the smallest resolution at which $x$ and $x'$ are clustered together $u_X(x,x') := \min \{ \delta\geq 0 \, | \, x\sim_{D_X(\delta)} x' \}$. We also consider the map $\Upsilon:\mathcal{U} \rightarrow \mathcal{D}$ constructed as follows: for a given ultrametric $u_X$ on the finite set $X$ and each $\delta \geq 0$ define the relation $\sim_{u_X(\delta)}$ on $X$ as $x \sim_{u_X(\delta)} x' \iff u_X(x,x')\leq \delta$. Further define $D_X(\delta) :=\big\{X \mod \sim_{u_X(\delta)}\big\}$ and $\Upsilon(X, u_X):= D_X$.

%
\begin{theorem}[\citealt{clust-um}]\label{theo_dendrograms_as_ultrametrics}
The maps $\Psi$ and $\Upsilon$ are both well defined. Furthermore, $\Psi\circ\Upsilon$ is the identity on $\mathcal{U}$ and $\Upsilon\circ\Psi$ is the identity on $\mathcal{D}$.
\end{theorem}

Given the equivalence between dendrograms and ultrametrics established by 
Theorem \ref{theo_dendrograms_as_ultrametrics} 
we can regard hierarchical clustering methods $\ccalH$ as inducing ultrametrics in node sets $X$ based on dissimilarity functions $A_X$. However, ultrametrics are particular cases of dissimilarity functions. Thus, we can reinterpret the method $\ccalH$ as a map [cf.~\eqref{eqn_clust_from_networks_to_dendrograms}]
\begin{equation}\label{eqn_clustering_from_networks_to_ultrametrics}
\ccalH:\ccalN\to\ccalU 
\end{equation}
mapping the space of networks $\ccalN$ to the space $\ccalU \subset \ccalN$ of networks endowed with ultrametrics.  For all $x, x' \in X$, the ultrametric value $u_X(x,x')$ induced by $\ccalH$ is the minimum resolution at which $x$ and $x'$ are co-clustered by $\ccalH$. Observe that the outcome of a hierarchical clustering method defines an ultrametric in the set $X$ even when the original data does not correspond to a metric, as is the case of asymmetric networks.
We say that two methods $\ccalH_1$ and $\ccalH_2$ are \emph{equivalent}, and we write $\ccalH_1 \equiv \ccalH_2$, if and only if $\ccalH_1(N) = \ccalH_2(N)$ for all $N \in \ccalN$.

\section {Axioms of value and transformation}\label{sec_axioms}

To study hierarchical clustering methods on asymmetric networks we start from intuitive notions that we translate into the axioms of value and transformation discussed in this section. 

\begin{figure}
  \centering
  \centerline{\def \thisplotscale {0.7}
\def \unit {\thisplotscale cm}
\def \xdendogram{{1, 2}}
\def \ydendogram{{1, 2}}

{\small
\begin{tikzpicture}[shorten >=2, scale = \thisplotscale]

    \node [blue vertex] at (-6.5,1) (p) {$p$};
    \node [blue vertex] at (-4,1) (q) {$q$};
    \path [-stealth](p) edge [bend left, above] node {$\alpha$} (q);	
    \path [-stealth] (q) edge [bend left, below] node {$\beta$}  (p);	
    
    \draw [-stealth] (-0.5,0) -- (5.3,0) node [below, at end] {$\delta$};
    \draw [-stealth] (0,-0.5) -- (0,2.9);
    
    \draw[thick] (0,0.7) -- ++(2.6,0) -- ++(0,1.2) -- +(-2.6,0) ++(0,-0.6) -- +(2.1,0);
    \draw[dashed](2.6,0.7) -- ++(0,-1.1) node [right, at end] {$\max(\alpha,\beta)$};
    \node [left] at (0,0.7) {$p$};
    \node [left] at (0,1.9) {$q$};
    
    \node at (-6,2.5) {$\vec{\Delta}_2(\alpha, \beta)$};
    \node at (-0.7,2.5) {$D_{p,q}$};
 
\end{tikzpicture}
}


    
}
  \vspace{-0.1in}
\caption{Axiom of Value. Nodes in a two-node network cluster at the minimum resolution at which both can influence each other.}
\vspace{-0.1in}
\label{fig_axioms_value_influence}
\end{figure}
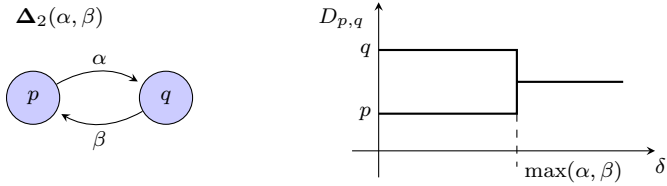

The Axiom of Value is obtained from considering the two-node network $\vec{\Delta}_2(\alpha, \beta)$ defined in Section~\ref{sec_preliminaries} and depicted in Fig. \ref{fig_axioms_value_influence}. We say that node $x$ is able to influence node $x'$ at resolution $\delta$ if the dissimilarity from $x$ to $x'$ is not greater than $\delta$. In two-node networks, our intuition dictates that a cluster is formed if nodes $p$ and $q$ are able to influence each other. This implies that the output dendrogram should be such that $p$ and $q$ are part of the same cluster at resolutions $\delta\geq\max(\alpha,\beta)$ that allow direct mutual influence. Conversely, we expect nodes $p$ and $q$ to be in separate clusters at resolutions $0 \leq \delta<\max(\alpha,\beta)$ that do {\it not} allow for mutual influence. At resolutions $\delta<\min(\alpha,\beta)$ there is no influence between the nodes and at resolutions $\min(\alpha,\beta) \leq \delta<\max(\alpha,\beta)$ there is unilateral influence from one node over the other. In either of the latter two cases the nodes are different in nature. If we think of dissimilarities as, e.g., trust, it means one node is trustworthy whereas the other is not. If we think of the network as a Markov chain, at resolutions $0 \leq \delta<\max(\alpha,\beta)$ the states are different singleton equivalence classes -- one of the states would be transient and the other one absorbent. Given that, according to \eqref{eqn_clustering_from_networks_to_ultrametrics}, a hierarchical clustering method is a map $\ccalH$ from networks to ultrametrics, we formalize this intuition as the following requirement on the set of admissible maps:

\myindentedparagraph{(A1) Axiom of Value} The ultrametric $(\{p,q\},u_{p,q})=\ccalH(\vec{\Delta}_2(\alpha, \beta))$ produced by $\ccalH$ applied to the two-node network $\vec{\Delta}_2(\alpha, \beta)$ satisfies $u_{p,q}(p,q) = \max(\alpha,\beta)$.

\vspace{0.075in}\noindent Clustering nodes $p$ and $q$ together at resolution $\delta=\max(\alpha,\beta)$ is somewhat arbitrary, as any monotone increasing function of $\max(\alpha, \beta)$ would be admissible. As a value claim, however, it means that the clustering resolution parameter $\delta$ is expressed in the same units as the elements of the dissimilarity function.

%

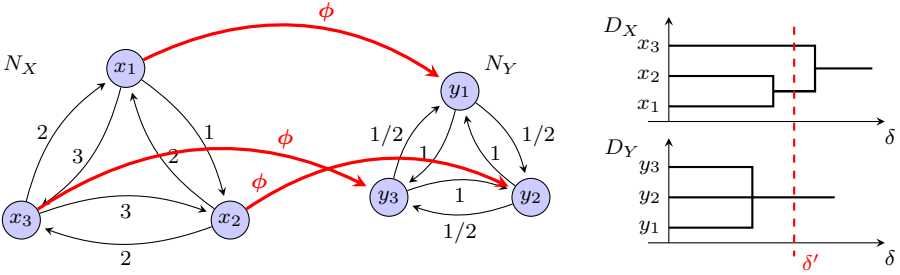
\begin{figure}
\centering
\def \thisplotscale {0.5}
\def \unit {\thisplotscale cm}

{\small
\begin{tikzpicture}[-stealth, shorten >=2, x = 1.1*\unit, y=1*\unit]

    \node [blue vertex] at (   0,  3.4) (1) {$x_1$};
    \node [blue vertex] at ( 2.5, -0.6) (2) {$x_2$};    
    \node [blue vertex] at (-2.5, -0.6) (3) {$x_3$};

    \path (1) edge [bend left=20, right] node {{$1$}} (2);	
    \path (2) edge [bend left=20, below] node {{$2$}} (3);
    \path (3) edge [bend left=20, left] node  {{$2$}} (1);    	

    \path (2) edge [bend left=20, right] node {{$2$}} (1);	
    \path (3) edge [bend left=20, below] node {{$3$}} (2);
    \path (1) edge [bend left=20, left]  node {{$3$}} (3);    	
    
    \node [blue vertex] at (8.0,2.8) (1p) {$y_1$};
    \node [blue vertex] at (9.7,0.0) (2p) {$y_2$};    
    \node [blue vertex] at (6.3,0.0) (3p) {$y_3$};

    \path (1p) edge [bend left=20, right] node {{ $1/2$}} (2p);	
    \path (2p) edge [bend left=20, below] node {{$1/2$}} (3p);
    \path (3p) edge [bend left=20, left]  node {{$1/2$}} (1p);    	

    \path (2p) edge [bend left=20, right] node {{$1$}} (1p);	
    \path (3p) edge [bend left=20, below] node {{$1$}} (2p);
    \path (1p) edge [bend left=20, left]  node {{$1$}} (3p);    	
    
    \path (1) edge [bend left, above, red, very thick, pos=0.6] node {$\bbphi$} (1p);	
    \path (2) edge [bend left, above, red, very thick, pos=0.05] node {$\bbphi$} (2p);	    
    \path (3) edge [bend left, above, red, very thick, pos=0.75] node {$\bbphi$} (3p);	    

   
    \draw [-stealth] (12.5,2) -- (18.3,2) node [below, at end] {$\delta$};
    \draw [-stealth] (13,2) -- (13,4.9);
    
    \draw[thick, -] (13,2.4) -- ++(2.5,0) -- ++(0,0.8) -- ++(-2.5,0) ++(0,0.8) -- ++(3.5,0) -- ++(0,-1.2) -- ++(-1,0) -- ++(1,0)--++ (0, 0.6)--++(1.5, 0);
    \node [left] at (13,2.4) {$x_1$};
    \node [left] at (13,3.2) {$x_2$};
    \node [left] at (13,4) {$x_3$};
    \node [left] at (12.5,4.5) {$D_{X}$};
    
    
    \draw [-stealth] (12.5,-1.2) -- (18.3,-1.2) node [below, at end] {$\delta$};
    \draw [-stealth] (13,-1.2) -- (13,1.7);
    
    \draw[thick, -] (13,-0.8) -- ++(2,0) -- ++(0,0.8) -- ++(-2,0) ++(0,0.8) -- ++(2,0) -- ++(0,-0.8) -- ++(2.1,0);
    \draw[dashed, thick, red, -](16,4.5) -- ++(0,-6.2) node [right, at end] {$\delta'$};
    \node [left] at (13,-0.8) {$y_1$};
    \node [left] at (13,0.) {$y_2$};
    \node [left] at (13,0.8) {$y_3$};
    \node [left] at (12.5,1.3) {$D_{Y}$};
     \node at (-2.5,3.5) {$N_{X}$};
      \node at (9, 3.5) {$N_{Y}$};

\end{tikzpicture}
}
\vspace{-0.2in}
\caption{Axiom of Transformation. If the network $N_X$ can be mapped to the network $N_Y$ using a dissimilarity-reducing map $\phi$, then for every resolution $\delta$ nodes clustered together in $D_X(\delta)$ must also be clustered in $D_Y(\delta)$. E.g., since points $x_1$ and $x_2$ are clustered together at resolution $\delta'$, their image through $\phi$, i.e. $y_1=\phi(x_1)$ and $y_2=\phi(x_2)$, must also be clustered together at this resolution.}
\vspace{-0.1in}
\label{fig_axiom_of_transformation}
\end{figure}

The second restriction on the space of allowable methods $\ccalH$ formalizes our expectations for the behavior of $\ccalH$ when confronted with a transformation of the underlying set $X$ and the dissimilarity function $A_X$; see Fig. \ref{fig_axiom_of_transformation}. Consider networks $N_X=(X,A_X)$ and $N_Y=(Y,A_Y)$ and denote by $D_X=\ccalH(X,A_X)$ and $D_Y=\ccalH(Y,A_Y)$ the corresponding dendrogram outputs. If we map all the nodes of the network $N_X=(X,A_X)$ into nodes of the network $N_Y=(Y,A_Y)$ in such a way that no pairwise dissimilarity is increased we expect the latter network to be more clustered than the former at any given resolution.  Intuitively, nodes in $N_Y$ are more capable of influencing each other, thus, clusters should be formed more easily. In terms of the respective dendrograms we expect that nodes co-clustered at resolution $\delta$ in $D_X$ are mapped to nodes that are also co-clustered at this resolution in $D_Y$. In order to formalize this notion, we introduce the concept of a \emph{dissimilarity-reducing map}. Given two networks $N_X=(X,A_X)$ and $N_Y=(Y,A_Y)$, map $\phi:X\to Y$ is {dissimilarity reducing} if it holds that $A_X(x,x')\geq A_Y(\phi(x),\phi(x'))$ for all $x,x'\in X$.

The Axiom of Transformation that we introduce next is a formal statement of the intuition described above:

\myindentedparagraph{(A2) Axiom of Transformation} Consider two networks $N_X=(X,A_X)$ and $N_Y=(Y,A_Y)$ and a dissimilarity-reducing map $\phi:X\to Y$, i.e. a map $\phi$ such that for all $x,x' \in X$ it holds that $A_X(x,x')\geq A_Y(\phi(x),\phi(x'))$. Then, for all $x, x' \in X$, the output ultrametrics $(X,u_X)=\ccalH(X,A_X)$ and $(Y,u_Y)=\ccalH(Y,A_Y)$ satisfy 
\begin{equation}\label{eqn_dissimilarity_reducing_ultrametric}
    u_X(x,x') \geq u_Y(\phi(x),\phi(x')).
\end{equation} 

\vspace{0.075in}\noindent We say that a hierarchical clustering method $\ccalH$ is {admissible} with respect to (A1) and (A2), or \emph{admissible} for short, if it satisfies axioms (A1) and (A2). 

For the particular case of symmetric networks $(X, A_X)$ we defined the single linkage dendrogram $\text{SL}_X$ through the equivalence relations in \eqref{eqn_single_linkage}. According to Theorem \ref{theo_dendrograms_as_ultrametrics} this dendrogram is equivalent to an ultrametric space that we denote by $(X, u^{\SL}_X)$. More specifically, as is well known \citep{clust-um}, the single linkage ultrametric $u^{\SL}_X$ in symmetric networks is given by 
\begin{align}\label{eqn_single_linkage_ultrametric}
   u^{\SL}_X(x,x') \ =\   \tdu^*_X(x, x') = \tdu^*_X(x', x) \ = \  \min_{C(x,x')} \,\,
                         \max_{i | x_i\in C(x,x')} A_X(x_i,x_{i+1}),
\end{align}
where we also used \eqref{eqn_nonreciprocal_chains} to write the last equality. 

%

\section{Influence modalities} \label{sec_axiomatic_structure}

The Axiom of Value states that, in order for two nodes to belong to the same cluster, they have to be able to exercise mutual influence on each other. When we consider a network with more than two nodes the concept of mutual influence is more difficult because it is possible to have direct influence as well as indirect chains of influence through other nodes. In this section we introduce two intuitive notions of mutual influence in networks of arbitrary size and show that they can be derived from the axioms of value and transformation. Besides their intrinsic value, these influence modalities are important for later developments in this paper; see, e.g. the proof of Theorem \ref{theo_extremal_ultrametrics}.

Consider first the intuitive notion that for two nodes to be part of a cluster there has to be a way for each of them to exercise influence on the other, either directly or indirectly. To formalize this idea, recall the concept of minimum loop cost \eqref{eqn_def_mlc}; see Fig. \ref{fig_axiom_of_influence}. For this network, the loops $[a,b,a]$ and $[b,a,b]$ have maximum cost $2$ corresponding to the link $(b,a)$ in both cases. All other two-node loops have cost $3$. All of the counterclockwise loops, e.g., $[a,c,b,a]$, have cost $3$ and any of the clockwise loops have cost $1$. Thus, the minimum loop cost of this network is $\mlc(X,A_X)=1$. 

For resolutions $0 \leq \delta<\mlc(X,A_X)$ it is impossible to find chains of mutual influence with maximum cost smaller than $\delta$ between any pair of points. Indeed, suppose we can link $x$ to $x'$ with a chain of maximum cost smaller than $\delta$, and also link $x'$ to $x$ with a chain having the same property. Then, we can form a loop with cost smaller than $\delta$ by concatenating these two chains. Thus, the intuitive notion that clusters cannot form at resolutions for which it is impossible to observe mutual influence can be translated into the requirement that no clusters can be formed at resolutions $\delta<\mlc(X,A_X)$. In terms of ultrametrics, this implies that it must be $u_X(x,x') \geq \mlc(X,A_X)$ for any $x \neq x' \in X$ as we formally state next:

%
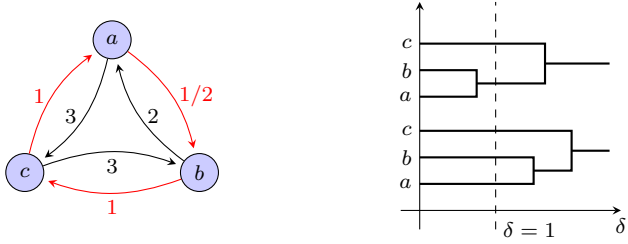
\begin{figure}
\centering
\def \thisplotscale {0.5}
\def \unit {\thisplotscale cm}

{\small
\begin{tikzpicture}[scale = \thisplotscale]

    \path   (-8.1, 3.5)   node [blue vertex] (1) {$a$}   
          ++( 2.3,-3.5) node [blue vertex] (2) {$b$}    
          ++(-4.6, 0)   node [blue vertex] (3) {$c$};

    \path (1) edge [bend left=20, red, right, shorten >=2, -stealth] node {$1/2$} (2);	
    \path (2) edge [bend left=20, red, below, shorten >=2, -stealth] node {$1$} (3);
    \path (3) edge [bend left=20, red, left, shorten >=2,  -stealth] node {$1$} (1);    	

    \path (2) edge [bend left=20, right, shorten >=2, -stealth] node {$2$} (1);	
    \path (3) edge [bend left=20, below, shorten >=2, -stealth] node {$3$} (2);
    \path (1) edge [bend left=20, left, shorten >=2, -stealth]  node {$3$} (3);

    \path [draw, -stealth]   (0,-1) ++ (-0.5,   0) -- ++ (5.8,0) node [below, at end] {$\delta$};
    \path [draw, -stealth]   (0,-1) ++ (   0,-0.5) -- ++ (0,6.0);

    \path [draw, thick]      (0,-1) ++ (0, 0.7) node [left] {$a$} -- ++(3.0,0) --
                                    ++ (0, 0.7) -- ++ (-3.0,0) node [left] {$b$};
    \path [draw, thick]      (0,-1) ++ (0, 2.1) node [left] {$c$} -- ++(4.0,0) --
                                    ++ (0,-1.05) -- ++ (-1,0);
    \path [draw, thick]      (0,-1) ++ (0, 1.575) ++ (4.0,0) -- ++(1,0);

    \path [draw, thick]      (0,1.3) ++ (0, 0.7) node [left] {$a$} -- ++(1.5,0) --
                                     ++ (0, 0.7) -- ++ (-1.5,0) node [left] {$b$};
    \path [draw, thick]      (0,1.3) ++ (0, 2.1) node [left] {$c$} -- ++(3.3,0) --
                                     ++ (0,-1.05) -- ++ (-1.8,0);
    \path [draw, thick]      (0,1.3) ++ (0, 1.575) ++ (3.3,0) -- ++(1.7,0);

    \path [draw, dashed]      (0,-1) ++ (2, -0.5) node [right] {$\delta=1$} -- ++(0,6);

\end{tikzpicture}
}
\vspace{-0.05in}
\caption{Property of Influence. No clusters can be generated at resolutions for which it is impossible to form influence loops. Here, the loop of minimum cost is formed by circling the network clockwise where the maximum cost encountered is $A_X(b,c)=A_X(c,a)=1$. The top dendrogram is invalid because $a$ and $b$ cluster at resolution $\delta<1$ whereas the bottom dendrogram satisfies the Property of Influence (P1).}
\vspace{-0.1in}
\label{fig_axiom_of_influence}
\end{figure} 

\myindentedparagraph{(P1) Property of Influence} For any network $N_X=(X,A_X)$ the ultrametric $(X, u_X)=\ccalH(X,A_X)$ is such that $u_X(x,x')$ for distinct nodes cannot be smaller than the minimum loop cost $\mlc(X,A_X)$ [cf. \eqref{eqn_def_mlc}] of the network, i.e. $u_X(x,x') \geq \mlc(X,A_X)$ for all $x \neq x'$.

\vspace{0.075in}
\noindent Since for the network in Fig. \ref{fig_axiom_of_influence} the minimum loop cost is $\mlc(X,A_X)=1$, then the Property of Influence implies that $u_X(x,x')\geq\mlc(X,A_X)=1$ for any pair of nodes $x \neq x'$. Equivalently, the output dendrogram is such that for resolutions $\delta<\mlc(X,A_X)=1$ each node is in its own block. Observe that (P1) does not imply that a cluster with more than one node {\it is} formed at resolution $\delta=\mlc(X,A_X)$ but states that achieving this minimum resolution is a necessary condition for the formation of clusters.

A second intuitive statement about influence in networks of arbitrary size comes in the form of the \emph{Extended Axiom of Value}. To introduce this concept define a family of canonical asymmetric networks $\vec{\Delta}_n(\alpha,\beta):=(\{1,\ldots,n\}, A_{n,\alpha,\beta})$, with $n\in\N$ and $\alpha, \beta > 0$, where the underlying node set $\{1,\ldots,n\}$ consists of the first $n$ natural numbers and the dissimilarity value $A_{n,\alpha,\beta}(i,j)$ between points $i$ and $j$ depends on whether $i>j$ or not. For points $i>j$ we let $A_{n,\alpha,\beta}(i,j)=\alpha$ whereas for points $i<j$ we have $A_{n,\alpha,\beta}(i,j)=\beta$. Recall that, by definition, $A_{n,\alpha,\beta}(i,i)=0$. In the network $\vec{\Delta}_n(\alpha,\beta)$ all pairs of nodes have dissimilarities $\alpha$ in one direction and $\beta$ in the other direction. This symmetry entails that all nodes should cluster together at the same resolution, and the requirement of mutual influence along with consistency with the Axiom of Value entails that this resolution should be $\max(\alpha,\beta)$. Before formalizing this definition notice that having clustering outcomes that depend on the ordering of the nodes in the space $\{1,\ldots,n\}$ is not desirable. Thus, we consider a permutation $\Pi= \{\pi_1, \pi_2, \ldots, \pi_n\}$ of $\{1,2, \ldots, n\}$ and the action $\Pi(A)$ on a dissimilarity function $A$, which we define by $\Pi(A)(i,j) = A(\pi_i,\pi_j)$ for all $i$ and $j$. Define now the network $\vec{\Delta}_n(\alpha,\beta,\Pi) : =(\{1,\ldots,n\},\Pi(A_{n,\alpha,\beta}))$. With this definition we can now formally introduce the Extended Axiom of Value as follows:

\myindentedparagraph{(A1') Extended Axiom of Value} Consider the network $\vec{\Delta}_n(\alpha,\beta,\Pi)=(\{1,\ldots,n\}, \\ \Pi(A_{n,\alpha,\beta}))$. Then, for all indices $n\in\mbN$, constants $\alpha,\beta>0$, and permutations $\Pi$ of $\{1,\ldots,n\}$, the outcome $(\{1,\ldots,n\},u)= \mathcal{H}\big(\vec{\Delta}_n(\alpha,\beta,\Pi)\big)$ satisfies $u(i,j) = \max(\alpha,\beta)$, for all pairs of nodes $i \neq j$.

\vspace{0.075in}
\noindent Observe that the Axiom of Value (A1) is subsumed into the Extended Axiom of Value for $n=2$. Further note that the minimum loop cost of $\vec{\Delta}_n(\alpha,\beta,\Pi)$ is $\max(\alpha, \beta)$. Combining this with the Property of Influence (P1), it follows that for the network $\vec{\Delta}_n(\alpha,\beta,\Pi)$ we must have $ u(i,j) \geq \mlc(\vec{\Delta}_n(\alpha,\beta))=\max(\alpha, \beta)$ for $i \neq j$. By the Extended Axiom of Value (A1') we have $u(i,j) = \max(\alpha,\beta)$ for $i \neq j$, which means that (A1') and (P1) are compatible requirements. We can then conceive of alternative axiomatic formulations where admissible methods are required to abide by the Axiom of Transformation (A2), the Property of Influence (P1), and either the (regular) Axiom of Value (A1) or the Extended Axiom of Value (A1') -- Axiom (A1) and (P1) are compatible because (A1) is a particular case of (A1') which we already argued is compatible with (P1). We will see in the following section that these two alternative axiomatic formulations are equivalent to each other in the sense that a clustering method satisfies one set of axioms if and only if it satisfies the other. We further show that (P1) and (A1') are implied by (A1) and (A2). As a consequence, it follows that both alternative axiomatic formulations are equivalent to simply requiring fulfillment of axioms (A1) and (A2).

\subsection{Equivalent axiomatic formulations}

We begin by stating the equivalence between admissibility with respect to (A1)-(A2) and (A1')-(A2). Furthermore, a theorem stating that methods admissible with respect to (A1) and (A2) satisfy the Property of Influence (P1) is presented next.

%
\begin{theorem}\label{theo_extended_value}
Assume the hierarchical clustering method $\ccalH$ satisfies the Axiom of Transformation (A2). Then, $\ccalH$ satisfies the Axiom of Value (A1) if and only if it satisfies the Extended Axiom of Value (A1').
\end{theorem}

The Extended Axiom of Value (A1') is stronger than the (regular) Axiom of Value (A1). However, Theorem \ref{theo_extended_value} shows that when considered together with the Axiom of Transformation (A2), both axioms of value are equivalent in the restrictions they impose in the set of admissible clustering methods $\ccalH$. In the following theorem we show that the Property of Influence (P1) can be derived from axioms (A1) and (A2).

\begin{theorem}\label{theo_influence_redundancy}
If a clustering method $\ccalH$ satisfies the Axiom of Value (A1) and the Axiom of Transformation (A2), then it satisfies the Property of Influence (P1).
\end{theorem}

The fact that (P1) is implied by (A1) and (A2) as claimed by Theorem \ref{theo_influence_redundancy} implies that adding (P1) as a third axiom on top of these two is moot. 
In the discussion leading to the introduction of the Axiom of Value (A1) in Section \ref{sec_axioms} we argued that the intuitive notion of a cluster dictates that it must be possible for co-clustered nodes to influence each other. In the discussion leading to the definition of the Property of Influence (P1) at the beginning of this section we argued that in networks with more than two nodes the natural extension is that co-clustered nodes must be able to influence each other either directly or through their indirect influence on other intermediate nodes. The Property of Influence is a codification of this intuition because it states the impossibility of cluster formation at resolutions where influence loops cannot be formed. While (P1) and (A1) seem quite different and seemingly independent, we have shown in this section that if a method satisfies axioms (A1) and (A2) it must satisfy (P1). Therefore, requiring direct influence on a two-node network as in (A1) restricts the mechanisms for indirect influence propagation so that clusters cannot be formed at resolutions that do not allow for mutual, possibly indirect, influence as stated in (P1). In that sense the restriction of indirect influence propagation in (P1) is not just {\it intuitively} reasonable but {\it formally} implied by the more straightforward restrictions on direct influence in (A1) and dissimilarity-reducing maps in (A2).

\begin{figure}
\centering
\def \thisplotscale {0.6}
\def \unit {\thisplotscale cm}

{
\begin{tikzpicture}[-stealth, shorten >=2 ,scale = \thisplotscale]

	\node[blue vertex] (x) {$x$} ;

	\path (x) ++ (4,0)   node [blue vertex]    (x1)   {$x_1$} 
	          ++ (4,0)   node [phantom vertex] (x2)   {$\ldots$}
	          ++ (0.5,0) node [phantom vertex] (xlm1) {$\ldots$} 
	          ++ (4,0)   node [blue vertex]    (xl)   {$x_{l-1}$}
	          ++ (4,0)    node [blue vertex]   (xp)   {$x'$}; 

	\path (x)    edge [bend left, above] node {$A_X(x,x_1)$}       (x1);	
	\path (x1)   edge [bend left, above] node {$A_X(x_1,x_2)$}     (x2);	
	\path (xlm1) edge [bend left, above] node {$A_X(x_{l-2},x_{l-1})$} (xl);	
	\path (xl)   edge [bend left, above] node {$A_X(x_{l-1},x')$}      (xp);	

	\path (x1)   edge [bend left, below] node {$A_X(x_1,x)$}       (x);
	\path (x2)   edge [bend left, below] node {$A_X(x_2,x_1)$}     (x1);
	\path (xl)   edge [bend left, below] node {$A_X(x_{l-1},x_{l-2})$} (xlm1);
	\path (xp)   edge [bend left, below] node {$A_X(x',x_{l-1})$}      (xl);

\end{tikzpicture}
} 
\vspace{-0.05in}
\caption{Reciprocal clustering. Nodes $x$ and $x'$ are clustered at resolution $\delta$ if they can be joined with a (reciprocal) chain whose maximum dissimilarity is smaller than or equal to $\delta$ in both directions [cf.~\eqref{eqn_reciprocal_clustering}].}
\vspace{-0.1in}
\label{fig_reciprocal_path}
\end{figure}
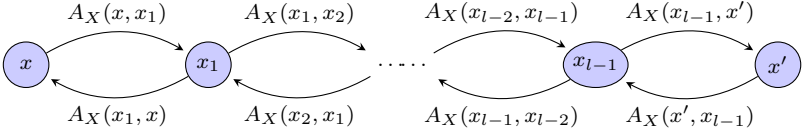

\section {Reciprocal and nonreciprocal clustering}\label{sec_reicprocal_and_nonreciprocal}

Pick any network $N_X=(X,A_X)\in\ccalN$. One particular clustering method satisfying axioms (A1)-(A2) can be constructed by considering the \emph{symmetric} dissimilarity
\begin{equation}\label{eqn_reciprocal_clustering_sym_matrix} 
 \bbarA_X(x, x') := \max(A_X(x,x'), A_X(x',x)),
\end{equation}
for all $x,x' \in X$. This effectively reduces the problem to clustering of symmetric data, a scenario in which single linkage clustering in \eqref{eqn_single_linkage} is known to satisfy axioms analogous to (A1)-(A2) \citep{clust-um}. Drawing upon this connection we define the \emph{reciprocal} clustering method $\ccalH^{\R}$ with output $(X,u^{\R}_X)=\ccalH^{\R}(X,A_X)$ as the one for which the ultrametric $u^{\R}_X(x,x')$ between points $x$ and $x'$ is given by
\begin{align}\label{eqn_reciprocal_clustering} 
    u^{\R}_X(x,x')
    &:= \min_{C(x,x')} \, \max_{i | x_i\in C(x,x')}
              \bbarA_X(x_i,x_{i+1}).
\end{align} 
An illustration of the definition in \eqref{eqn_reciprocal_clustering} is shown in Fig. \ref{fig_reciprocal_path}. We search for chains $C(x,x')$ linking nodes $x$ and $x'$. For a given chain we walk from $x$ to $x'$  and for every link, connecting say $x_i$ with $x_{i+1}$, we determine the maximum dissimilarity in both directions, i.e. the value of $\bbarA_X(x_i, x_{i+1})$. We then determine the maximum across all the links in the chain. The reciprocal ultrametric $u^{\R}_X(x,x')$ between points $x$ and $x'$ is the minimum of this value across all possible chains. Recalling the equivalence of dendrograms and ultrametrics provided by Theorem \ref{theo_dendrograms_as_ultrametrics}, we know that $\R_X$, the dendrogram produced by reciprocal clustering, clusters $x$ and $x'$ together for resolutions $\delta\geq u^{\R}_X(x,x')$. Combining the latter observation with \eqref{eqn_reciprocal_clustering}, we can write the reciprocal clustering equivalence classes as
\begin{equation}\label{eqn_reciprocal_clustering_dendrogram} 
   x\sim_{\R_X(\delta)} x' \iff  
       \min_{C(x,x')} \, \max_{i | x_i\in C(x,x')}\bbarA_X(x_i,x_{i+1})\leq\delta.
\end{equation}
Comparing \eqref{eqn_reciprocal_clustering_dendrogram} with the definition of single linkage in \eqref{eqn_single_linkage} with $\tdu^*_X(x, x')$ as defined in \eqref{eqn_nonreciprocal_chains}, we see that reciprocal clustering is equivalent to single linkage for the symmetrized network $N=(X,\bbarA_X)$ where dissimilarities between nodes are symmetrized to the maximum value of each directed dissimilarity.

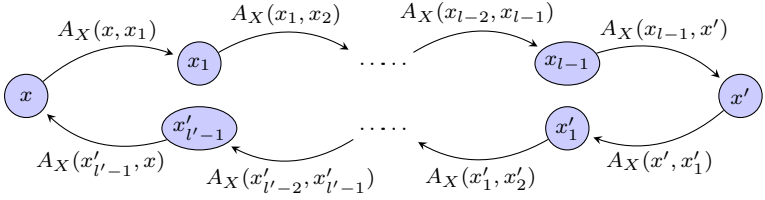
\begin{figure*}
\centering
\def \thisplotscale {0.57}
\def \unit {\thisplotscale cm}

{\small
\begin{tikzpicture}[-stealth, shorten >=2 ,scale = \thisplotscale]
	
	\node[blue vertex] (x) {$x$} ;

	\path (x) ++ (4,0.75)   node [blue vertex]    (x1)   {$x_1$} 
	          ++ (4,0)   node [phantom vertex] (x2)   {$\ldots$}
	          ++ (0.5,0) node [phantom vertex] (xlm1) {$\ldots$} 
	          ++ (4,0)   node [blue vertex]    (xl)   {$x_{l-1}$}
	          ++ (4,-0.75)  node [blue vertex]    (xp)   {$x'$}; 

	\path (x) ++ (4,-0.75)  node [blue vertex]    (xlp)   {$x'_{l'-1}$} 
	          ++ (4,0)   node [phantom vertex] (xlm1p) {$\ldots$}
	          ++ (0.5,0) node [phantom vertex] (x2p)   {$\ldots$} 
	          ++ (4,0)   node [blue vertex]    (x1p)   {$x'_{1}$}; 

	\path (x)    edge [bend left, above] node {$A_X(x,x_1)$}       (x1);	
	\path (x1)   edge [bend left, above] node {$A_X(x_1,x_2)$}     (x2);	
	\path (xlm1) edge [bend left, above] node {$A_X(x_{l-2},x_{l-1})$} (xl);	
	\path (xl)   edge [bend left, above] node {$A_X(x_{l-1},x')$}      (xp);	

	\path (xp)    edge [bend left, below] node {$A_X(x',x'_1)$}       (x1p);
	\path (x1p)   edge [bend left, below] node {$A_X(x'_1,x'_2)$}     (x2p);
	\path (xlm1p) edge [bend left, below] node {$A_X(x'_{l'-2},x'_{l'-1})$} (xlp);
	\path (xlp)   edge [bend left, below] node {$A_X(x'_{l'-1},x)$}       (x);

\end{tikzpicture}
} 
\vspace{-0.05in}
\caption{Nonreciprocal clustering. Nodes $x$ and $x'$ are clustered at resolution $\delta$ if they can be joined in both directions with possibly different chains of maximum dissimilarity not greater than $\delta$ [cf.~\eqref{eqn_nonreciprocal_clustering}].}
\vspace{-0.1in}
\label{fig_nonreciprocal_path}
\end{figure*}

For the method $\ccalH^{\R}$ specified in \eqref{eqn_reciprocal_clustering} to be a properly defined hierarchical clustering method, we need to establish that $u^{\R}_X$ is a valid ultrametric. It is clear that $u^{\R}_X(x,x')=0$ only if $x=x'$ and that $u^{\R}_X(x,x')=u^{\R}_X(x',x)$ because the definition is symmetric on $x$ and $x'$. To verify that the strong triangle inequality in \eqref{eqn_strong_triangle_inequality} holds, let $C^*(x,x')$ and $C^*(x',x'')$ be chains that achieve the minimum in \eqref{eqn_reciprocal_clustering} for $u^{\R}_X(x,x')$ and $u^{\R}_X(x',x'')$, respectively. The maximum cost in the concatenated chain $C(x,x'')=C^*(x,x')\uplus C^*(x',x'')$ does not exceed the maximum cost in each individual chain. Thus, while the cost may be smaller on a different chain, the chain $C(x,x'')$ suffices to bound $u^{\R}_X(x,x'') \leq \max \big( u^{\R}_X(x,x'), u^{\R}_X(x',x'')\big)$ as in \eqref{eqn_strong_triangle_inequality}. It is also possible to prove that $\ccalH^{\R}$ satisfies axioms (A1)-(A2), as we do next.

\begin{proposition}\label{prop_reciprocal_axioms}
The reciprocal clustering method $\ccalH^{\R}$ is valid and admissible. I.e., $u_X^{\R}$ in \eqref{eqn_reciprocal_clustering} is an ultrametric for all networks and $\ccalH^{\R}$ satisfies axioms (A1)-(A2).
\end{proposition}
\begin{myproofnoname}
That $u_X^{\R}$ conforms to the definition of an ultrametric was proved in the paragraph preceding this proposition. To see that the Axiom of Value (A1) is satisfied, pick an arbitrary two-node network $\vec{\Delta}_2(\alpha, \beta)$ as defined in Section \ref{sec_preliminaries} and denote by $(\{p,q\}, u^{\R}_{p,q})=\ccalH^\R(\vec{\Delta}_2(\alpha, \beta))$ the output of applying the reciprocal clustering method to $\vec{\Delta}_2(\alpha, \beta)$. Since every possible chain from $p$ to $q$ must contain $p$ and $q$ as consecutive nodes, applying the definition in \eqref{eqn_reciprocal_clustering} yields $u^{\R}_{p,q}(p,q) = \max \big(A_{p,q}(p,q), \\ A_{p,q}(q,p) \big) = \max(\alpha,\beta)$. Axiom (A1) is thereby satisfied. 

To show fulfillment of Axiom (A2), consider two networks $(X, A_X)$ and $(Y, A_Y)$, a dissimilarity-reducing map $\phi:X \to Y$ and define $(X, u^\R_X):=\ccalH^\R(X, A_X)$ and $(Y, u^\R_Y):=\ccalH^\R(Y, A_Y)$. For an arbitrary pair of nodes $x, x' \in X$, denote by $C^*_X(x,x')=[x=x_0,\ldots, x_l=x']$ a chain that achieves the minimum reciprocal cost in \eqref{eqn_reciprocal_clustering} so as to write $u^{\R}_X(x,x') = \max_{i | x_i\in C^*_X(x,x')} \, \bbarA_X(x_i,x_{i+1})$. Consider the transformed chain $C_Y(\phi(x),\phi(x'))=[\phi(x)=\phi(x_0),\ldots, \phi(x_l)=\phi(x')]$ in the set $Y$. Since the transformation $\phi$ does not increase dissimilarities we have that for all links in this chain $A_Y(\phi(x_i),\phi(x_{i+1}))\leq A_X(x_i,x_{i+1})$ and $A_Y(\phi(x_{i+1}),\phi(x_i))\leq A_X(x_{i+1},x_i)$. This implies that
\begin{align}\label{eqn_theo_reciprocal_axioms_pf_50}
    \max_{i | \phi(x_i)\in C_Y(\phi(x),\phi(x'))} \bbarA_Y(\phi(x_i),\phi(x_{i+1})) \leq u^{\R}_X(x,x').
\end{align}
Further note that $C_Y(\phi(x),\phi(x'))$ is a particular chain joining $\phi(x)$ and $\phi(x')$ whereas the reciprocal ultrametric is the minimum across all such chains. Therefore,
\begin{equation}\label{eqn_theo_reciprocal_axioms_pf_60}
    u^{\R}_Y(\phi(x),\phi(x')) \leq \max_{i | \phi(x_i)\in C_Y(\phi(x),\phi(x'))} \bbarA_Y(\phi(x_i),\phi(x_{i+1})).
   \end{equation}
Substituting \eqref{eqn_theo_reciprocal_axioms_pf_50} in \eqref{eqn_theo_reciprocal_axioms_pf_60}, the fulfillment of Axiom (A2) follows.
\end{myproofnoname}

In reciprocal clustering, nodes $x$ and $x'$ belong to the same cluster at a resolution $\delta$ whenever we can go back and forth from $x$ to $x'$ at a maximum cost $\delta$ through the same chain. By contrast, in \emph{nonreciprocal} clustering we relax the restriction about the chain being the same in both directions and cluster nodes $x$ and $x'$ together if there are chains, possibly different, linking $x$ to $x'$ and $x'$ to $x$. To state this definition in terms of ultrametrics consider a given network $N=(X,A_X)$ and recall the definition of the unidirectional minimum chain cost $\tdu^*_X$ in \eqref{eqn_nonreciprocal_chains}. We define the {nonreciprocal} clustering method $\ccalH^{\NR}$ with output $(X,u^{\NR}_X)=\ccalH^{\NR}(X,A_X)$ as the one for which the ultrametric $u^{\NR}_X(x,x')$ between points $x$ and $x'$ is given by the maximum of the unidirectional minimum chain costs $\tdu^*_X(x, x')$ and $\tdu^*_X(x', x)$ in each direction,
\begin{equation}\label{eqn_nonreciprocal_clustering} 
    u^{\NR}_X(x,x') := \max \Big( \tdu^*_X(x,x'),\ \tdu^*_X(x',x )\Big).
\end{equation} 
An illustration of the definition in \eqref{eqn_nonreciprocal_clustering} is shown in Fig. \ref{fig_nonreciprocal_path}. We consider forward chains $C(x,x')$ going from $x$ to $x'$ and backward chains $C(x',x)$ going from $x'$ to $x$. For each of these chains we determine the maximum dissimilarity across all the links in the chain. We then search independently for the best forward chain $C(x,x')$ and the best backward chain $C(x',x)$ that minimize the respective maximum dissimilarities across all possible chains. The nonreciprocal ultrametric $u^{\NR}_X(x,x')$ between points $x$ and $x'$ is the maximum of these two minimum values.

As it is the case with reciprocal clustering we can verify that $u_X^{\NR}$ is a properly defined ultrametric and that, as a consequence, the nonreciprocal clustering method $\ccalH^{\NR}$ is properly defined. Identity and symmetry are immediate. For the strong triangle inequality consider chains $C^*(x,x')$ and $C^*(x',x'')$ that achieve the minimum costs in $\tdu^*_X(x,x')$  and $\tdu^*_X(x',x'')$ as well as the chains $C^*(x'',x')$ and $C^*(x',x)$ that achieve the minimum costs in $\tdu^*_X(x'',x')$ and $\tdu^*_X(x',x)$. The concatenation of these chains permits concluding that $u^{\NR}_X(x,x'')\leq \max \big( u^{\NR}_X(x,x'), u^{\NR}_X(x',x'')\big)$, which is the strong triangle inequality in \eqref{eqn_strong_triangle_inequality}. The method $\ccalH^{\NR}$ also satisfies axioms (A1)-(A2) as the following proposition states.

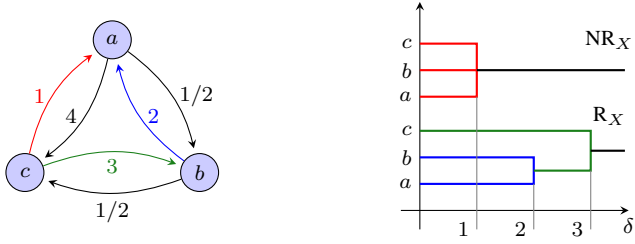
\begin{figure}
\centering
\def \thisplotscale {0.5}
\def \unit {\thisplotscale cm}

{\small
\begin{tikzpicture}[scale = \thisplotscale]

    \path   (-8.1, 3.5)   node [blue vertex] (1) {$a$}   
          ++( 2.3,-3.5) node [blue vertex] (2) {$b$}    
          ++(-4.6, 0)   node [blue vertex] (3) {$c$};

    \path (1) edge [bend left=20, right, shorten >=2, -stealth] node {$1/2$} (2);	
    \path (2) edge [bend left=20, below, shorten >=2, -stealth] node {$1/2$} (3);
    \path (3) edge [bend left=20, red, left, shorten >=2,  -stealth] node {$1$} (1);    	

    \path (2) edge [bend left=20, blue, right, shorten >=2, -stealth] node {$2$} (1);	
    \path (3) edge [bend left=20, mygreen, below, shorten >=2, -stealth] node {$3$} (2);
    \path (1) edge [bend left=20, left, shorten >=2, -stealth]  node {$4$} (3);

    \path [draw, -stealth]   (0,-1) ++ (-0.5,   0) -- ++ (6.0,0) node [below, at end] {$\delta$};
    \path [draw, -stealth]   (0,-1) ++ (   0,-0.5) -- ++ (0,6.0);

    \path [draw, thick, draw=blue]     (0,-1) ++ (0, 0.7) node [left] {$a$} -- ++(3.0,0) --
                                          ++ (0, 0.7) -- ++ (-3.0,0) node [left] {$b$};
    \path [draw, thick, draw=mygreen]  (0,-1) ++ (0, 2.1) node [left] {$c$} -- ++(4.5,0) --
                                          ++ (0,-1.05) -- ++ (-1.5,0);
    \path [draw, thick]                (0,-1) ++ (0, 1.575) ++ (4.5,0) -- ++(0.9,0);

    \path [draw, thick, draw=red] (0,1.3) ++ (0, 0.7) node [left] {$a$} -- ++(1.5,0) --
                                     ++ (0, 0.7) -- ++ (-1.5,0) node [left] {$b$};
    \path [draw, thick, draw=red] (0,1.3) ++ (0, 2.1) node [left] {$c$} -- ++(1.5,0) --
                                     ++ (0,-0.7);
    \path [draw, thick]           (0,1.3) ++ (0, 1.4) ++ (1.5,0) -- ++(3.9,0);


    \path [draw=black!50] (0,-1) ++ (3,   -0.5) node [left] {$2$} -- ++(0,0.5) -- ++(0,0.7);
    \path [draw=black!50] (0,-1) ++ (4.5, -0.5) node [left] {$3$} -- ++(0,0.5) -- ++(0,1.05);
    \path [draw=black!50] (0,-1) ++ (1.5, -0.5) node [left] {$1$} -- ++(0,0.5) -- ++(0,3.0);
    
    \node at (5,1.5) {$\R_X$};
        \node at (5,3.5) {$\NR_X$};

\end{tikzpicture}
}
\vspace{-0.05in}
\caption{Reciprocal and nonreciprocal dendrograms. An example network with its corresponding reciprocal (bottom) and nonreciprocal (top) dendrograms.}
\vspace{-0.1in}
\label{fig_reciprocal_nonreciprocal_dendrograms}
\end{figure}

\begin{proposition}\label{prop_nonreciprocal_axioms}
The nonreciprocal clustering method $\ccalH^{\NR}$ is valid and admissible. I.e., $\!u_X^{\NR}\!$ in \eqref{eqn_nonreciprocal_clustering} is an ultrametric for all networks and $\ccalH^{\NR}\!$ satisfies axioms (A1)-\!(A2).
\end{proposition}
\begin{myproofnoname} 
That $\ccalH^{\NR}$ outputs valid ultrametrics was already argued prior to the statement of Proposition \ref{prop_nonreciprocal_axioms}. The proof for admissibility of $\ccalH^{\NR}$ is omitted since it is analogous to that of admissibility of $\ccalH^{\R}$ (cf. Theorem~\ref{prop_reciprocal_axioms}).
\end{myproofnoname}


The reciprocal and nonreciprocal dendrograms for an example network are shown in Fig. \ref{fig_reciprocal_nonreciprocal_dendrograms}. Notice that these dendrograms are \emph{different}. In the reciprocal dendrogram nodes $a$ and $b$ cluster together at resolution $\delta=2$ due to their direct connections $A_X(a,b)=1/2$ and $A_X(b,a)=2$. Node $c$ joins this cluster at resolution $\delta=3$ because it links bidirectionally with $b$ through the chain $[b,c]$ whose maximum cost is $A_X(c,b)=3$.  The optimal reciprocal chain linking $a$ and $c$ is $[a,b,c]$ whose maximum cost is also $A_X(c,b)=3$. In the nonreciprocal dendrogram we can link nodes with different chains in each direction. As a consequence, $a$ and $b$ cluster together at resolution $\delta=1$ because the directed cost of the chain $[a,b]$ is $A_X(a,b)=1/2$ and the directed cost of the chain $[b,c,a]$ is $A_X(c,a)=1$. Similar chains demonstrate that $a$ and $c$ as well as $b$ and $c$ also cluster together at resolution $\delta=1$.

\section {Extremal ultrametrics}\label{sec_extremal_ultrametrics}

Given that we have constructed two admissible methods satisfying axioms (A1)-(A2), the question whether these two constructions are the only possible ones arises and, if not, whether they are special in some sense. We prove in this section that reciprocal and nonreciprocal clustering are a peculiar pair in that all possible admissible clustering methods are contained between them in a well-defined sense. To explain this sense properly, observe that since reciprocal chains [cf. Fig. \ref{fig_reciprocal_path}] are particular cases of nonreciprocal chains [cf. Fig. \ref{fig_nonreciprocal_path}] we must have that $u^{\NR}_X(x,x') \leq  u^{\R}_X(x,x')$ for all pairs of nodes $x,x'$. I.e., nonreciprocal ultrametrics do not exceed reciprocal ultrametrics. An important characterization is that any method $\ccalH$ satisfying axioms (A1)-(A2) yields ultrametrics that lie between $u_X^{\NR}$ and $u_X^{\R}$ as we formally state next.

%
\begin{theorem}\label{theo_extremal_ultrametrics}
Consider an admissible clustering method $\ccalH$ satisfying axioms (A1)-(A2). For an arbitrary given network $N=(X,A_X)$ denote by $(X,u_X)=\ccalH(N)$ the output of $\ccalH$ applied to $N$. Then, for all pairs of nodes $x,x' \in X$
\begin{equation}\label{eqn_theo_extremal_ultrametrics} 
    u^{\NR}_X(x,x') \leq  u_X(x,x') \leq  u^{\R}_X(x,x'),
\end{equation} 
where $u^{\NR}_X$ and $u^{\R}_X$ denote the nonreciprocal and reciprocal ultrametrics as defined by \eqref{eqn_nonreciprocal_clustering} and \eqref{eqn_reciprocal_clustering}, respectively.
\end{theorem}


%
According to Theorem \ref{theo_extremal_ultrametrics}, nonreciprocal clustering applied to a given network $N=(X,A_X)$ yields a uniformly minimal ultrametric among those output by all clustering methods satisfying axioms (A1)-(A2). Reciprocal clustering yields a uniformly maximal ultrametric. Any other clustering method abiding by (A1)-(A2) yields an ultrametric such that the value $u_X(x,x')$ for any two points $x, x' \in X$ lies between the values $u^{\NR}_X(x,x')$ and $u^{\R}_X(x,x')$ assigned by nonreciprocal and reciprocal clustering. In terms of dendrograms, \eqref{eqn_theo_extremal_ultrametrics} implies that among all possible clustering methods, the smallest possible resolution at which nodes are clustered together is the one corresponding to nonreciprocal clustering. The highest possible resolution is the one that corresponds to reciprocal clustering.

\subsection{Hierarchical clustering on symmetric networks}\label{secsymmetric_networks}

Restrict attention to the subspace $\ccalM \subset \ccalN$ of symmetric networks, that is $N=(X,A_X)\in\mathcal{M}$ if and only if $A_X(x,x')=A_X(x',x)$ for all $x, x' \in X$. When restricted to the space $\ccalM$ reciprocal and nonreciprocal clustering are equivalent methods because, for any pair of points, minimizing nonreciprocal chains are always reciprocal -- more precisely there may be multiple minimizing nonreciprocal chains but at least one of them is reciprocal. To see this formally, first fix $x,x'\in X$ and observe that in symmetric networks the symmetrization in \eqref{eqn_reciprocal_clustering_sym_matrix} is unnecessary because  $\bbarA_X(x_i,x_{i+1}) = A_X(x_i,x_{i+1}) = A_X(x_{i+1},x_i)$ and the definition of reciprocal clustering in \eqref{eqn_reciprocal_clustering}  reduces to
\begin{align}\label{eqn_reciprocal_symetric_networks} 
    u^{\R}_X(x,x') \! = \!\! \min_{C(x,x')} \, \max_{i | x_i\in C(x,x')} \!\!\!\! A_X(x_i,x_{i+1}) \! = \!\! \min_{C(x',x)} \, \max_{i | x_i\in C(x',x)} \!\!\!\! A_X(x_i,x_{i+1}).
\end{align} 
Further note that the costs of any given chain $C(x,x')=[x=x_0, x_1, \ldots , x_{l-1}, x_l=x']$ and its reciprocal $C(x',x)=[x'=x_l, x_{l-1}, \ldots , x_{1}, x_0=x]$ are the same. It follows that directed minimum chain costs $\tdu^*_X(x, x')=\tdu^*_X(x', x)$ are equal and according to \eqref{eqn_nonreciprocal_clustering} equal to the nonreciprocal ultrametric 
\begin{align}\label{eqn_nonreciprocal_symetric_networks}
   u^{\NR}_X(x, x')= \tdu^*_X(x, x') = \tdu^*_X(x', x) = u^{\R}_X(x,x').
\end{align} 
To write the last equality in \eqref{eqn_nonreciprocal_symetric_networks} we used the definitions   of $\tdu^*_X(x, x')$ and $\tdu^*_X(x', x)$ in \eqref{eqn_nonreciprocal_chains} which are correspondingly equivalent to the first and second equality in \eqref{eqn_reciprocal_symetric_networks}.

By further comparison of the ultrametric definition of single linkage in \eqref{eqn_single_linkage_ultrametric} with \eqref{eqn_nonreciprocal_symetric_networks} the equivalence of reciprocal, nonreciprocal, and single linkage clustering in symmetric networks follows
\begin{align}\label{eqn_nonreciprocal_sl_reciprocal}
   u^{\NR}_X(x, x')= u^{\SL}_X(x,x') = u^{\R}_X(x,x').
\end{align} 
The equivalence in \eqref{eqn_nonreciprocal_symetric_networks} along with Theorem \ref{theo_extremal_ultrametrics} demonstrates that when considering the application of hierarchical clustering methods $\ccalH:\ccalM\to\ccalU$ to symmetric networks, there exist a unique method satisfying (A1)-(A2). The equivalence in \eqref{eqn_nonreciprocal_sl_reciprocal} shows that this method is single linkage. Before stating this result formally let us define the symmetric version of the Axiom of Value:

\myindentedparagraph{(B1) Symmetric Axiom of Value} Consider a symmetric two-node network $\vec{\Delta}_2(\alpha, \alpha)$. The ultrametric $(\{p, q\},u_{p,q})=\ccalH(\vec{\Delta}_2(\alpha, \alpha))$ satisfies $u_{p,q}(p,q) = \alpha$.

\vspace{0.075in}\noindent Since there is only one dissimilarity in a symmetric network with two nodes, (B1) states that they cluster together at the resolution that connects them to each other. We can now prove that single linkage is the unique hierarchical clustering method in symmetric networks that is admissible with respect to (B1) and (A2). 

%
\begin{corollary}\label{cor_single_linkage} Let $\ccalH:\ccalM\to\ccalU$ be a hierarchical clustering method for symmetric networks and $\ccalH^{\SL}$ be the single linkage method with output ultrametrics as defined in \eqref{eqn_single_linkage_ultrametric}. If $\ccalH$ satisfies axioms (B1) and (A2) then $\ccalH\equiv\ccalH^{\SL}$.
\end{corollary}
\begin{myproofnoname} When restricted to symmetric networks, (B1) and (A1) are equivalent statements. Thus, $\ccalH$ satisfies the hypotheses of Theorem \ref{theo_extremal_ultrametrics} and, as a consequence, \eqref{eqn_theo_extremal_ultrametrics} is true for any pair of points $x,x'$ of any network $N\in\ccalM$. But by \eqref{eqn_nonreciprocal_sl_reciprocal} nonreciprocal, single linkage, and reciprocal ultrametrics coincide. Thus, we can reduce \eqref{eqn_theo_extremal_ultrametrics} to $u^{\text{SL}}_X(x,x') \leq  u_X(x,x') \leq  u^{\text{SL}}_X(x,x')$, implying that $\ccalH\equiv\ccalH^{\SL}$. \end{myproofnoname}

The uniqueness result claimed by Corollary \ref{cor_single_linkage} strengthens the uniqueness result by \citet[Theorem 18]{clust-um}. To explain the differences consider the symmetric version of the Property of Influence. In a symmetric network there is always a loop of minimum cost of the form $[x,x',x]$ for some pair of points $x,x'$. Indeed, say that $C^*(x^*,x^*)$ is one of the loops achieving the minimum cost in \eqref{eqn_def_mlc} and let $A_X(x,x')=\mlc(X,A_X)$ be the maximum dissimilarity in this loop. Then, the cost of the loop $[x,x',x]$ is $A_X(x,x')=A_X(x',x)=\mlc(X,A_X)$ which means that either the loop $C^*(x^*,x^*)$ was already of the form $[x,x',x]$ or that the cost of the loop $[x,x',x]$ is the same as $C^*(x^*,x^*)$. In any event, there is a loop of minimum cost of the form $[x,x',x]$ which implies that in symmetric networks we must have
\begin{equation}\label{eqn_mlc_equals_sep_symetric}
    \mlc(X,A_X) = \min_{x \neq x'} A_X(x, x') = \sep(X,A_X),
\end{equation}
We can introduce the symmetric version of the Property of Influence:

\myindentedparagraph{(Q1) Symmetric Property of Influence} For any symmetric network $N_X=(X,A_X)$ the output $(X, u_X)=\ccalH(N_X)$ is such that $u_X(x,x')$ for distinct points cannot be smaller than the network separation, i.e. $u_X(x,x') \geq \sep(N_X)$ for all $x \neq x'$.

\vspace{0.075in}\noindent \citet{clust-um} define admissibility with respect to (B1), (A2), and (Q1), which corresponds to conditions (I), (II), and (III) of their Theorem~18. Corollary \ref{cor_single_linkage} shows that Property (Q1) is redundant when given axioms (B1) and (A2) -- respectively, Condition (III) by \citet[Theorem 18]{clust-um} is redundant when given conditions (I) and (II). Corollary \ref{cor_single_linkage} also shows that single linkage is the unique admissible method for all symmetric, not necessarily metric, networks.

%
\section{Alternative axiomatic constructions}\label{sec_alternative_axioms}
The axiomatic framework that we adopted allows alternative constructions by modifying the underlying set of axioms. Among the axioms in Section \ref{sec_axioms}, the Axiom of Value (A1) is perhaps the most open to interpretation. Although we required the two-node network in Fig. \ref{fig_axioms_value_influence} to first cluster into one single block at resolution $\max(\alpha,\beta)$ corresponding to the largest dissimilarity and argued that this was reasonable in most situations, it is also reasonable to accept that in some situations the two nodes should be clustered together as long as one of them is able to influence the other. To account for this possibility we replace the Axiom of Value by the following alternative.

\myindentedparagraph{(A1'') Alternative Axiom of Value} The ultrametric $(\{p,q\},u_{p,q}):=\ccalH(\vec{\Delta}_2(\alpha, \beta))$ output by $\ccalH$ from the two-node network $\vec{\Delta}_2(\alpha, \beta)$ satisfies $u_{p,q}\!(p,q)\! = \!\min(\alpha,\beta)$.

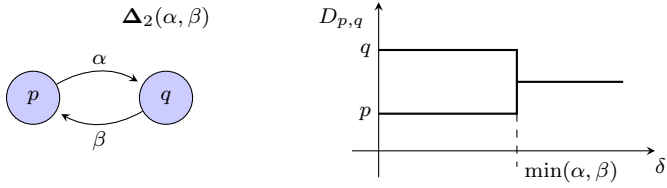
\begin{figure}
  \centering
  \centerline{\def \thisplotscale {0.7}
\def \unit {\thisplotscale cm}
\def \xdendogram{{1, 2}}
\def \ydendogram{{1, 2}}

{\small
\begin{tikzpicture}[shorten >=2, scale = \thisplotscale]

    \node [blue vertex] at (-6.5,1) (p) {$p$};
    \node [blue vertex] at (-4,1) (q) {$q$};
    \path [-stealth](p) edge [bend left, above] node {$\alpha$} (q);	
    \path [-stealth] (q) edge [bend left, below] node {$\beta$}  (p);	
    
    \draw [-stealth] (-0.5,0) -- (5.3,0) node [below, at end] {$\delta$};
    \draw [-stealth] (0,-0.5) -- (0,2.9);
    
    \draw[thick] (0,0.7) -- ++(2.6,0) -- ++(0,1.2) -- +(-2.6,0) ++(0,-0.6) -- +(2.1,0);
    \draw[dashed](2.6,0.7) -- ++(0,-1.1) node [right, at end] {$\min(\alpha,\beta)$};
    \node [left] at (0,0.7) {$p$};
    \node [left] at (0,1.9) {$q$};
    
        \node at (-4,2.5) {$\vec{\Delta}_2(\alpha, \beta)$};
    \node at (-0.7,2.5) {$D_{p,q}$};
 
\end{tikzpicture}
}


    
}
\vspace{-0.1in}
\caption{Alternative Axiom of Value. For a two-node network, nodes are clustered together at the minimum resolution at which one of them can influence the other.}
\vspace{-0.1in}
\label{fig_two_node_network_dendrogram_modif}
\end{figure}

\vspace{0.075in}\noindent Axiom (A1'') replaces the requirement of bidirectional influence in Axiom (A1) to unidirectional influence; see Fig. \ref{fig_two_node_network_dendrogram_modif}. We say that a clustering method $\ccalH$ is admissible with respect to the alternative axioms if it satisfies axioms (A1'') and (A2).

The property of influence (P1), which is a keystone in the proof of Theorem \ref{theo_extremal_ultrametrics}, is not compatible with the Alternative Axiom of Value (A1''). Indeed, just observe that the minimum loop cost of the two-node network in Fig. \ref{fig_two_node_network_dendrogram_modif} is $\mlc(\vec{\Delta}_2(\alpha, \beta))=\max(\alpha, \beta)$ whereas in (A1'') we are requiring the output ultrametric to be $u_{p,q}(p,q)=\min(\alpha,\beta)$. We therefore have that Axiom (A1'') itself implies $u_{p,q}(p,q) = \min(\alpha, \beta) \\ < \max(\alpha, \beta) = \mlc(\vec{\Delta}_2(\alpha, \beta))$ for the cases when $\alpha \neq \beta$. Thus, we reformulate (P1) into the Alternative Property of Influence (P1') that we define next.

\myindentedparagraph{(P1') Alternative Property of Influence} For any network $N_X\!=\!(X,A_X)$ the output ultrametric $(X, u_X)\!=\!\ccalH(N_X)$ is such that $u_X(x,x')$ for distinct points cannot be smaller than the separation of the network, $u_X(x,x')\! \geq \!\sep(N_X)$ for all~$x~\neq~x'$.

\vspace{0.075in}\noindent Observe that the Alternative Property of Influence (P1') coincides with the Symmetric Property of Influence (Q1) defined in Section \ref{secsymmetric_networks}. This is not surprising because for symmetric networks the Axiom of Value (A1) and the Alternative Axiom of Value (A1'') impose identical restrictions. Moreover, since the separation of a network cannot be larger than its minimum loop cost, the Alternative Property of Influence (P1') is implied by the (regular) Property of Influence (P1), but not vice versa.

The Alternative Property of Influence (P1') states that no clusters are formed at resolutions at which there are no unidirectional influences between any pair of nodes and is consistent with the Alternative Axiom of Value (A1''). Moreover, in studying methods admissible with respect to (A1'') and (A2), (P1') plays a role akin to the one played by (P1) when studying methods that are admissible with respect to (A1) and (A2). In particular, as (P1) is implied by (A1) and (A2), (P1') is true if (A1'') and (A2) hold as we assert in the following theorem.

%
\begin{theorem}\label{theo_alternative_influence}
If a clustering method $\ccalH$ satisfies the Alternative Axiom of Value (A1'') and the Axiom of Transformation (A2) then it also satisfies the Alternative Property of Influence (P1').
\end{theorem}

Theorem \ref{theo_alternative_influence} admits the following interpretation. In (A1'') we require two-node networks to cluster at the resolution where unidirectional influence occurs. When we consider (A1'') in conjunction with (A2) we can translate this requirement into a statement about clustering in arbitrary networks. Such requirement is the Alternative Property of Influence (P1') which prevents nodes to cluster at resolutions at which no influence exists between \emph{any} two nodes.

\subsection{Unilateral clustering}\label{sec_unilateral_clustering}

Mimicking the developments in Sections \ref{sec_axioms}-\ref{sec_extremal_ultrametrics}, we move on to identify and define methods that satisfy axioms (A1'')-(A2) and then bound the range of admissible methods with respect to these axioms. To do so, let $N=(X, A_X)$ be a given network and consider the dissimilarity function $\hat{A}_X(x, x') : = \min(A_X(x, x'),A_X(x', x))$, for all $x, x' \in X$. Notice that, as opposed to the definition of $\bar{A}_X$, where the symmetrization is done by means of a $\max$ operation, $\hat{A}$ is defined by using a $\min$ operation.
We define the \emph{unilateral} clustering method $\ccalH^\U$ with output ultrametric $(X, u^\U_X)= \ccalH^\U(N)$, where $u^\U_X$ is defined as
\begin{equation}\label{eqn_unilateral_clustering_2} 
    u^{\U}_X(x,x') :=
       \min_{C(x,x')} \, \max_{i | x_i\in C(x,x')} \hat{A}_X(x_i,x_{i+1}),
\end{equation} 
for all $x, x' \in X$.
To show that $\ccalH^\U$ is a properly defined clustering method, we need to establish that $u_X^\U$ as defined in \eqref{eqn_unilateral_clustering_2} is a valid ultrametric. However, comparing \eqref{eqn_unilateral_clustering_2} and \eqref{eqn_single_linkage_ultrametric} we see that $\ccalH^\U(X, A_X) \equiv \ccalH^{\SL}(X, \hat{A}_X)$, i.e. applying the unilateral clustering method to an asymmetric network $(X, A_X)$ is equivalent to applying single linkage clustering method to the symmetrized network $(X, \hat{A}_X)$. Since we know that single linkage produces a valid ultrametric when applied to any symmetric network such as $(X, \hat{A}_X)$, \eqref{eqn_unilateral_clustering_2} is a properly defined ultrametric. Furthermore, it can be shown that $\ccalH^\U$ satisfies axioms (A1'') and (A2).

\begin{proposition}\label{prop_unilateral_axioms}
The unilateral clustering method $\ccalH^\U$ with output ultrametrics defined in \eqref{eqn_unilateral_clustering_2} satisfies axioms (A1'') and (A2). 
\end{proposition}

In the case of admissibility with respect to (A1) and (A2), nonreciprocal and reciprocal clustering are two different admissible methods which bound every other possible clustering method satisfying (A1)-(A2) (cf. Theorem \ref{theo_extremal_ultrametrics}). In contrast, in the case of admissibility with respect to (A1'') and (A2), unilateral clustering is the \emph{unique} admissible method as stated in the following theorem.

\begin{theorem}\label{theo_unilateral_unicity}
Let $\ccalH$ be a hierarchical clustering method satisfying axioms (A1'') and (A2). Then, $\ccalH \equiv \ccalH^\U$ where $\ccalH^\U$ is the unilateral clustering.
\end{theorem}

By Theorem \ref{theo_unilateral_unicity}, the space of methods that satisfy the Alternative Axiom of Value (A1'') and the Axiom of Transformation (A2) is inherently simpler than the space of methods that satisfy the (regular) Axiom of value (A1) and the Axiom of Transformation (A2). Further note that in the case of symmetric networks, for all $x, x' \in X$ we have $\hat{A}_X(x, x')=A_X(x, x')=A_X(x', x)$ and as a consequence unilateral clustering is equivalent to single linkage as it follows from comparison of \eqref{eqn_single_linkage_ultrametric} and 
\eqref{eqn_unilateral_clustering_2}. Thus, the result in Theorem \ref{theo_unilateral_unicity} reduces to the statement in Corollary \ref{cor_single_linkage}, which was derived upon observing that in symmetric networks reciprocal and nonreciprocal clustering yield identical outcomes. The fact that reciprocal, nonreciprocal, and unilateral clustering all coalesce into single linkage when restricted to symmetric networks is consistent with the fact that the Axiom of Value (A1) and the Alternative Axiom of Value (A1'') are both equivalent to the Symmetric Axiom of Value (B1) when restricted to symmetric dissimilarities.

\subsection{Agnostic Axiom of Value}\label{sec_agnostic_axiom_of_value}

Axiom (A1) stipulates that every two-node network  $\vec{\Delta}_2(\alpha, \beta)$ is clustered into a single block at resolution $\max(\alpha,\beta)$, whereas Axiom (A1'') stipulates that they should be clustered at $\min(\alpha,\beta)$. One can also be agnostic with respect to this issue and say that both of these situations are admissible. An agnostic version of axioms (A1) and (A1'') is given next.

\myindentedparagraph{(A1\emph{'''}) Agnostic Axiom of Value} The ultrametric $(X,u_{p,q})=\ccalH(\vec{\Delta}_2(\alpha, \beta))$ produced by $\ccalH$ applied to the two-node network $\vec{\Delta}_2(\alpha, \beta)$ satisfies $\min(\alpha,\beta) \leq u_X(p,q) \leq \max(\alpha,\beta)$.

\vspace{0.075in}
\noindent Since fulfillment of (A1) or (A1'') implies fulfillment of (A1'''), any admissible clustering method with respect to the original axioms (A1)-(A2) or with respect to the alternative axioms (A1'')-(A2) must be admissible with respect to the agnostic axioms (A1''')-(A2). In this sense, (A1''')-(A2) is the most general combination of axioms described in this paper. For methods that are admissible with respect to (A1''') and (A2) we can bound the range of outcome ultrametrics as stated next.

\begin{theorem}\label{theo_extremal_ultrametrics_2}
Consider a clustering method $\ccalH$ satisfying axioms (A1\emph{'''}) and (A2). For an arbitrary given network $N=(X,A_X)$ denote by $(X, u_X)=\ccalH(X,A_X) $ the outcome of $\ccalH$ applied to $N$. Then, for all pairs of nodes $x,x' \in X$
\begin{equation}\label{eqn_theo_extremal_ultrametrics_2} 
    u^{\U}_X(x,x') \leq  u_X(x,x') \leq  u^{\R}_X(x,x'),
\end{equation} 
where $u^{\U}_X(x,x')$ and $u^{\R}_X(x,x')$ denote the unilateral and reciprocal ultrametrics as defined by \eqref{eqn_unilateral_clustering_2} and \eqref{eqn_reciprocal_clustering}, respectively.
\end{theorem}

By Theorem \ref{theo_extremal_ultrametrics_2}, given an asymmetric network $(X,A_X)$, any hierarchical clustering method abiding by axioms (A1''') and (A2) produces outputs contained between those corresponding to  two methods. The first method, unilateral clustering, symmetrizes $A_X$ by calculating $\hat{A}_X(x, x')=\min(A_X(x, x'), A_X(x',x))$ for all $x, x' \in X$ and computes single linkage on $(X,\hat{A}_X)$. The other method, reciprocal clustering, symmetrizes $A_X$ by calculating $\bbarA_X(x, x')=\max(A_X(x, x'), A_X(x', x))$ for all $x, x' \in X$ and computes single linkage on $(X,\bbarA_X)$.

%
\section{Conclusions}\label{sec_conclusions}

We presented an axiomatic construction of hierarchical clustering for asymmetric networks. Even though the notion of proximity between nodes -- hence, the concept of clustering -- is unclear when we are given directed dissimilarities, we determined desirable properties that clustering methods should satisfy. These properties were translated into the axioms of value and transformation. We then presented two clustering methods -- reciprocal and nonreciprocal -- that abide by these axioms. In reciprocal clustering, node clusters are formed based on path of bidirectional influence whereas in nonreciprocal clustering the influence in both directions can be propagated via different paths. More interestingly, we showed that any other method satisfying both axioms must be contained between reciprocal and nonreciprocal clustering in a well-defined sense. We also analyzed alternative axiomatic constructions. The construction based on the Extended Axiom of Value, though seemingly stronger, was shown to be equivalent to the original axiomatic framework. A different construction, based on an Alternative Axiom of Value, gave rise to a unique admissible clustering method, named unilateral clustering where unidirectional influence is sufficient for the formation of clusters. Finally, when applied to symmetric networks, all hierarchical clustering methods considered here boil down to single linkage, in which case the characterization results presented generalize and expand existing results for clustering of finite metric spaces.

\section{Appendix: Proofs}\label{sec_appendix}

\begin{myproof}[of Theorem \ref{theo_extended_value}]
In proving Theorem \ref{theo_extended_value}, we make use of the following lemma.

%
\begin{lemma}\label{lemma_axiom_redundancy}
Let $N=(X, A_X)$ be any network and $\delta$ any positive constant. Suppose that $x,x' \in X$ are such that their associated minimum chain cost [cf. \eqref{eqn_nonreciprocal_chains}] satisfies $\tdu^*_X(x, x') \geq \delta$. Then, there exists a partition $P_\delta(x,x')=\{B_\delta(x), B_\delta(x')\}$ of the node set $X$ into blocks $B_\delta(x)$ and $B_\delta(x')$ with $x \in B_\delta(x)$ and $x' \in B_\delta(x')$ such that $A_X(b, b') \geq \delta$, for all points $b \in B_\delta(x)$ and $b' \in B_\delta(x')$.
\end{lemma}

%
\begin{myproofnoname}
We prove this by contradiction. If a partition $P_\delta(x,x')=\{B_\delta(x), B_\delta(x')\}$ with $x \in B_\delta(x)$ and $x' \in B_\delta(x)$ satisfying Lemma~\ref{lemma_axiom_redundancy} does not exist for all pairs of points $x,x'\in X$ satisfying $\tdu^*_X(x, x') \geq \delta$, then there is at least one pair of nodes $x,x'\in X$ satisfying $\tdu^*_X(x, x') \geq \delta$ such that for {\it all} partitions of $X$ into two blocks $P=\{B, B'\}$ with $x \in B$ and $x' \in B'$ we can find at least a pair of elements $b_P \in B$ and $b'_P \in B'$ for which
\begin{equation}\label{eqn_lem_axiom_redundancy_03} 
   A_X(b_P, b'_P) < \delta.
\end{equation}
Begin by considering the partition $P_1=\{B_1, B'_1\}$ where $B_1 = \{ x \}$ and $B'_1 = X \backslash \{x\}$. Since \eqref{eqn_lem_axiom_redundancy_03} is true for all partitions having $x\in B$ and $x'\in B'$ and $x$ is the unique element of $B_1$, there must exist a node $b'_{P_1} \in B'_1$ such that  
\begin{equation}\label{eqn_lem_axiom_redundancy_04} 
   A_X(x, b'_{P_1}) <  \delta. 
\end{equation}
Hence, the chain $C(x, b'_{P_1})= [x, b'_{P_1}]$ composed of these two nodes has cost smaller than $\delta$. Moreover, since $\tdu^*_X(x, b'_{P_1})$ represents the minimum cost among all chains $C(x, b'_{P_1})$ linking $x$ to $b'_{P_1}$, we can assert that $\tdu^*_X(x, b'_{P_1}) \leq  A_X(x, b'_{P_1}) <  \delta$. Consider now the partition $P_2=\{B_2, B'_2\}$ where $B_2= \{ x, b'_{P_1} \}$ and $B'_2=X \backslash B_2$. From \eqref{eqn_lem_axiom_redundancy_03}, there must exist a node $b'_{P_2} \in B'_2$ that satisfies at least one of the two following conditions: i) $A_X(x, b'_{P_2}) <  \delta$, or ii) $A_X(b'_{P_1}, b'_{P_2}) <  \delta$. If i) is true, the chain $C(x, b'_{P_2})=[x, b'_{P_2}]$ has cost smaller than $\delta$. If ii) is true, we combine the dissimilarity bound  with the one in \eqref{eqn_lem_axiom_redundancy_04} to conclude that the chain $C(x, b'_{P_2})=[x, b'_{P_1}, b'_{P_2}]$ has cost smaller than $\delta$. In either case we conclude that there exists a chain $C(x, b'_{P_2})$ linking $x$ to $b'_{P_2}$ whose cost is smaller than $\delta$. Therefore, the minimum chain cost must satisfy $\tdu^*_X(x, b'_{P_2}) <  \delta$. We can repeat this process iteratively where, e.g., partition $P_3$ is composed by $B_3= \{ x, b'_{P_1}, b'_{P_2} \}$ and $B'_3=X\backslash B_3$, to obtain partitions $P_1, P_2, ... , P_{n-1}$ and corresponding nodes $b'_{P_1}, b'_{P_2}, \dots, b'_{P_{n-1}}$ such that the associated minimum chain cost satisfies $\tdu^*_X(x, b'_{P_i}) <  \delta$, for all $i$. Observe that nodes $b'_{P_i}$ are distinct by construction and distinct from $x$. Since there are $n$ nodes in the network it must be that $x'=b'_{P_k}$ for some $i \in \{1, \ldots , n-1\}$, entailing that $\tdu^*_X(x, x') <  \delta$, and reaching a contradiction. \end{myproofnoname}

%

\noindent
Continuing with the proof of Theorem \ref{theo_extended_value}, to show that (A1)-(A2) imply (A1')-(A2) let $\ccalH$ be a method that satisfies (A1) and (A2) and denote by $(\{1, 2, \ldots, n\},u_{n, \alpha, \beta})= \ccalH(\vec{\Delta}_n(\alpha,\beta,\Pi))$. We want to prove that (A1') is satisfied which means that we have to show that for all indices $n\in\mbN$, constants $\alpha,\beta>0$, permutations $\Pi$ of $\{1,\ldots,n\}$, and points $i\neq j$, we have $u_{n, \alpha, \beta}(i,j)=\max(\alpha,\beta)$. We will do so by showing both i) $u_{n, \alpha, \beta}(i,j)\ \leq\ \max(\alpha,\beta)$, and ii) $u_{n, \alpha, \beta}(i,j)\ \geq\ \max(\alpha,\beta)$, for all $n\in\mbN$, $\alpha,\beta>0$, $\Pi$, and $i\neq j$.

To prove i), define the two-node network $N_\mathrm{max}:= \vec{\Delta}_2(\max(\alpha, \beta), \max(\alpha, \beta))$ and define $\big(\{p, q\}, u_{p,q}\big) := \ccalH(N_\mathrm{max})$. Since $\ccalH$ abides by (A1),
\begin{align}\label{eqn_lem_axiom_redundancy_110}
   u_{p,q}(p,q)
      = \max\big(\max(\alpha, \beta),\max(\alpha, \beta)\big)
      = \max(\alpha, \beta).
\end{align}
Consider now the map $\phi_{i,j}:\{p,q\} \to \{1,\ldots,n\}$ from $N_\mathrm{max}$ to the permuted canonical network $\vec{\Delta}_n(\alpha,\beta,\Pi)$ where $\phi_{i,j}(p)=i$ and $\phi_{i,j}(q)=j$. Since dissimilarities in $\vec{\Delta}_n(\alpha,\beta,\Pi)$ are either $\alpha$ or $\beta$ and the dissimilarities in the two-node network are $\max(\alpha,\beta)$ it follows that the map $\phi_{i,j}$ is dissimilarity reducing regardless of the particular values of $i$ and $j$. Since the method $\ccalH$ was assumed to satisfy (A2) as well, we must have $u_{p,q} (p,q) \geq  u_{n, \alpha, \beta}\big(\phi_{i,j}(p),\phi_{i,j}(q)\big) = u_{n, \alpha, \beta}(i,j)$. Inequality i) follows form substituting \eqref{eqn_lem_axiom_redundancy_110} into this last expression.

In order to show inequality ii), pick two arbitrary distinct nodes $i, j \in \{1,\ldots,n\}$ in the node set of $\vec{\Delta}_n(\alpha,\beta,\Pi)$. Denote by $C(i,j)$ and $C(j,i)$ two minimizing chains in the definition  \eqref{eqn_nonreciprocal_chains} of the directed minimum chain costs $\tdu^*_{n, \alpha, \beta}(i, j)$ and $\tdu^*_{n, \alpha, \beta}(j, i)$ respectively. Observe that at least one of the following two inequalities must be true $\tdu^*_{n, \alpha, \beta}(i, j) \geq \max(\alpha, \beta)$ or $\tdu^*_{n, \alpha, \beta}(j, i) \geq \max(\alpha, \beta)$. Indeed, if both inequalities were false, the concatenation of $C(i,j)$ and $C(j,i)$ would form a loop $C(i,i)=C(i,j) \uplus C(j,i)$ of cost strictly less than $\max(\alpha, \beta)$. This cannot be true because $\max(\alpha, \beta)$ is the minimum loop cost of the network $\vec{\Delta}_n(\alpha,\beta,\Pi)$.

Without loss of generality assume $\tdu^*_{n, \alpha, \beta}(i, j) \geq \max(\alpha, \beta)$ is true and consider $\delta = \max(\alpha, \beta)$. By Lemma \ref{lemma_axiom_redundancy} we are therefore guaranteed to find a partition of the node set $\{1,\ldots,n\}$ into two blocks $B_\delta(i)$ and $B_\delta(j)$  with $i \in B_\delta(i)$ and $j \in B_\delta(j)$ such that for all $b \in B_\delta(i)$ and $b' \in B_\delta(j)$ it holds that 
\begin{equation}\label{eqn_b_bprime_max}
   \Pi(A_{n,\alpha,\beta})(b, b') \geq \delta = \max(\alpha, \beta).
\end{equation}
Define a two-node network $N_\mathrm{min}:=\vec{\Delta}_2(\max(\alpha, \beta), \min(\alpha, \beta))=(\{r, s\}, A_{r,s})$ where $A_{r,s}(r,s)=\max(\alpha, \beta)$ and $A_{r,s}(s,r)=\min(\alpha, \beta)$ and define $(\{r,s\}, u_{r,s}) \\ := \ccalH(N_\mathrm{min})$. Since the method $\ccalH$ satisfies (A1) we must have
\begin{equation}\label{eqn_max_max_min_a_b}
   u_{r,s}(r, s)
      = \max \big( \max(\alpha, \beta), \min(\alpha, \beta) \big) 
      = \max(\alpha, \beta).
\end{equation}
Consider the map $\phi'_{i,j} : \{1,\ldots,n\} \to \{r, s\}$ such that $\phi'_{i,j}(b)=r$ for all $b \in B_\delta(i)$ and $\phi'_{i,j}(b')=s$ for all $b' \in B_\delta(j)$. The map $\phi'_{i,j}$ is dissimilarity reducing because
\begin{equation}\label{eqn_phi_dissim_reducing_i_j}
      \Pi(A_{n,\alpha,\beta})(k,l) \geq A_{r,s}(\phi'_{i,j}(k), \phi'_{i,j}(l)),
\end{equation}
for all $k, l \in \{1,\ldots,n\}$. To see the validity of \eqref{eqn_phi_dissim_reducing_i_j} consider three different possible cases. If $k$ and $l$ belong both to the same block, i.e., either $k,l \in B_\delta(i)$ or $k,l \in B_\delta(j)$, then $\phi'_{i,j}(k)=\phi'_{i,j}(l)$ and $A_{r,s}(\phi'_{i,j}(k), \phi'_{i,j}(l))=0$, immediately satisfying \eqref{eqn_phi_dissim_reducing_i_j}. If $k \in B_\delta(j)$ and $l \in B_\delta(i)$ it holds that $A_{r,s}(\phi'_{i,j}(k), \phi'_{i,j}(l)) = A_{r,s}(s, r)= \min(\alpha, \beta)$ which cannot exceed $\Pi(A_{n,\alpha,\beta})(k,l)$ which is either equal to $\alpha$ or $\beta$. If $k \in B_\delta(i)$ and $l \in B_\delta(j)$, then we have $A_{r,s}(\phi'_{i,j}(k), \phi'_{i,j}(l)) = A_{r,s}(r, s)= \max(\alpha, \beta)$ but we also have $\Pi(A_{n,\alpha,\beta})(k,l)= \max(\alpha, \beta)$ as it follows by taking $b=k$ and $b'=l$ in \eqref{eqn_b_bprime_max}, thus, again satisfying \eqref{eqn_phi_dissim_reducing_i_j}.

Since $\ccalH$ fulfills the Axiom of Transformation (A2) we must have
\begin{equation}\label{eqn_phi_dissim_reducing_i_j_2}
   u_{n, \alpha, \beta}(i,j) \geq u_{r,s}\big(\phi'_{i,j}(i), \phi'_{i,j}(j)\big) = u_{r,s}(r, s).
\end{equation}
Substituting \eqref{eqn_max_max_min_a_b} in \eqref{eqn_phi_dissim_reducing_i_j_2} we obtain the inequality ii). Combining both inequalities i) and ii), it follows that $u_{n, \alpha, \beta}(i,j)= \max(\alpha,\beta)$. Thus, admissibility with respect to (A1)-(A2) implies admissibility with respect to (A1')-(A2). The opposite implication is immediate since (A1) is a particular case of (A1'), concluding the proof.\end{myproof}

\begin{myproof}[of Theorem~\ref{theo_influence_redundancy}]
We show that if a clustering method satisfies axioms (A1') and (A2) then it satisfies the Property of Influence (P1). Notice that this result, combined with Theorem~\ref{theo_extended_value}, implies the statement of Theorem~\ref{theo_influence_redundancy}.
The following lemma is instrumental in the ensuing proof.

%
\begin{lemma}\label{lemma_influence_redundancy}
Let $N=(X, A_X)$ be an arbitrary network with $n$ nodes and $\vec{\Delta}_n(\alpha,\beta)=(\{1,\ldots,n\}, A_{n,\alpha,\beta})$ be the canonical network with $0 < \alpha \leq \sep(X, A_X)$ and $\beta=\mlc(X, A_X)$. Then, there exists a bijective dissimilarity-reducing map $\phi:X\to\{1,\ldots,n\}$, i.e. $A_X(x, x') \geq A_{n, \alpha,\beta}(\phi(x), \phi(x'))$, for all $x, x' \in X$.
 \end{lemma}

%
\begin{myproofnoname} To construct the map $\phi$ consider the function $P:X \to \mathcal{P}(X)$ from the node set $X$ to its power set $\mathcal{P}(X)$ such that $P(x) :=\{ x' \in X \, | \, x' \neq x \,\, , \,\, A_X(x', x)<\beta\}$, for all $x \in X$.
Having $r \in P(s)$ for some $r,s \in X$ implies that $A_X(r, s) < \beta=\mlc(X,A_X)$. An important observation is that we must have a node $x\in X$ whose $P$-image is empty. Otherwise, pick a node $x_n\in X$ and construct the chain $[x_0, x_1, \ldots , x_n]$ where the $i$th element $x_{i-1}$ of the chain is in the $P$-image of $x_i$. From the definition of $P$ it follows that all dissimilarities along this chain satisfy $A_X(x_{i-1},x_{i})<\beta=\mlc(X,A_X)$. But since the chain $[x_0, x_1, \ldots , x_n]$ contains $n+1$ elements, at least one node must be repeated. Hence, we have found a loop for which all dissimilarities are bounded above by $\beta=\mlc(X,A_X)$, which is impossible because it contradicts the definition of the minimum loop cost in \eqref{eqn_def_mlc}.  We can then find a node $x_{i_1}$ for which $P(x_{i_1})=\emptyset$. Fix $\phi(x_{i_1})=1$. 

Select now a node $x_{i_2}\neq x_{i_1}$ whose $P$-image is either $\{x_{i_1}\}$ or $\emptyset$, which we write jointly as $P(x_{i_2}) \subseteq \{x_{i_1}\}$. Following a similar reasoning to the previous one, such a node must exist and fix $\phi(x_{i_2})=2$. Repeat this process $k$ times so that at step $k$ we have $\phi(x_{i_k})=k$ for a node $x_{i_k} \not\in \{x_{i_1},x_{i_2},\ldots x_{i_{k-1}}\}$ whose P-image is a subset of the nodes already picked, i.e., $P(x_{i_k}) \subseteq \{x_{i_1}, \ldots x_{i_{k-1}}\}$. 
Since all the nodes $x_{i_k}$ are different, the map $\phi$ with $\phi(x_{i_k})=k$ is bijective. By construction, $\phi$ is such that for all $l>k$, $x_{i_l}\notin P(x_{i_k})$. From the definition of $P$, this implies that the dissimilarity from $x_{i_l}$ to $x_{i_k}$ must satisfy $A_X(x_{i_l},x_{i_k}) \geq \beta$, for all $l>k$. Moreover, from the definition of the canonical matrix $A_{n, \alpha,\beta}$ we have that $A_{n, \alpha,\beta}(\phi(x_{i_l}), \phi(x_{i_k})) = A_{n, \alpha,\beta}(l,k) = \beta$ for all $l>k$. By combining these two expressions, we conclude that $A_X(x, x') \geq A_{n, \alpha,\beta}(\phi(x), \phi(x'))$ is true for all points with $\phi(x)>\phi(x')$. When $\phi(x)<\phi(x')$, we have $A_{n, \alpha,\beta}(\phi(x), \phi(x'))=\alpha$ which was assumed to be bounded above by the separation of the network $(X, A_X)$, thus, $A_{n, \alpha,\beta}(\phi(x), \phi(x'))$ is not greater than any positive dissimilarity in the range of $A_X$.
\end{myproofnoname}

%
\noindent
Continuing the main proof of Theorem \ref{theo_influence_redundancy}, consider a given arbitrary network $N=(X, A_X)$ with $X=\{x_1, x_2, ... , x_n\}$ and define $(X, u_X) :=\ccalH(X,A_X)$. The method $\ccalH$ is known to satisfy (A1') and (A2) and we want to show that it satisfies (P1) for which we need to show that $u_X(x,x')\geq\mlc(X,A_X)$ for all $x\neq x'$. 

Consider the canonical network $\vec{\Delta}_n(\alpha,\beta)=(\{1,\ldots,n\}, A_{n,\alpha,\beta})$ with $\beta=\mlc(X, A_X)$ being the minimum loop cost of the network $N$ and $\alpha>0$ a constant not exceeding the separation of the network. Thus, we have $\alpha\leq\sep(X,A_X)\leq\mlc(X,A_X)=\beta$. Note that networks $N$ and $\vec{\Delta}_n(\alpha,\beta)$ have equal number of nodes.

%
Defining $(\{1,\ldots,n\}, u_{\alpha,\beta}) := \ccalH(\vec{\Delta}_n(\alpha,\beta))$, since $\ccalH$ satisfies the Extended Axiom of Value (A1'), then for all indices $i, j \in \{1,\ldots,n\}$ with $i \neq j$ we have
\begin{equation}\label{eqn_ultram_epsilon_m_1}
   u_{\alpha,\beta}(i,j) = \max(\alpha,\beta) = \beta = \mlc(X,A_X) .
\end{equation}
Further, focus on the bijective dissimilarity-reducing map considered in Lemma \ref{lemma_influence_redundancy} and notice that since $\ccalH$ satisfies (A2) it follows that for all $x, x' \in X$ 
\begin{equation}\label{eqn_ultram_epsilon_m_2}
u_X(x,x') \geq u_{\alpha,\beta}(\phi(x), \phi(x')).
\end{equation}
Since the equality in \eqref{eqn_ultram_epsilon_m_1} is true for all $i\neq j$ and since all points $x\neq x'$ are mapped to points $\phi(x)\neq\phi(x')$ because $\phi$ is bijective, \eqref{eqn_ultram_epsilon_m_2} implies $u_X(x,x') \geq \beta = \mlc(X, A_X)$, for all distinct $x, x' \in X$.
\end{myproof}

\begin{myproof}[of Theorem~\ref{theo_extremal_ultrametrics}] We prove the theorem by showing both inequalities in \eqref{eqn_theo_extremal_ultrametrics}.
\begin{myproof}[of ${\bf u^{\NR}_X(x,x') \leq  u_X(x,x')}$] Recall that validity of (A1)-(A2) implies validity of (P1) by Theorem~\ref{theo_influence_redundancy}. Consider the nonreciprocal clustering equivalence relation $\sim_{\NR_X(\delta)}$ at resolution $\delta$ according to which $x \sim_{\NR_X(\delta)} x'$ if and only if $x$ and $x'$ belong to the same nonreciprocal cluster at resolution $\delta$. Notice that this is true if and only if $u^{\NR}_X(x,x')\leq\delta$. Further consider the set $Z := X \mod \sim_{\NR_X(\delta)}$ of corresponding equivalence classes and the map $\phi_{\delta}:X\to Z$ that maps each point of $X$ to its equivalence class. Notice that $x$ and $x'$ are mapped to the same point $z$ if they belong to the same cluster at resolution $\delta$.

We define the network $N_Z:=(Z,A_Z)$ by endowing $Z$ with the dissimilarity $A_Z$ derived from the dissimilarity $A_X$ as
\begin{equation}\label{eqn_theo_extremal_ultrametrics_pf_020}
    A_Z(z,z') := \min_{x\in\phi_\delta^{-1}(z), x'\in\phi_\delta^{-1}(z')} A_X(x,x').
\end{equation} 
The dissimilarity $A_Z(z,z')$ compares all the dissimilarities $A_X(x,x')$ between a member of the equivalence class $z$ and a member of the equivalence class $z'$ and sets $A_Z(z,z')$ to the value corresponding to the least dissimilar pair; see Fig. \ref{fig_proof_theo_extremal_ultrametrics}. Notice that according to construction, the map $\phi_\delta$ is dissimilarity reducing $A_X(x,x') \geq A_Z(\phi_\delta(x),\phi_\delta(x'))$, because we either have $A_Z(\phi_\delta(x),\phi_\delta(x'))=0$ if $x$ and $x'$ are co-clustered at resolution $\delta$, or $A_X(x,x') \geq \min_{x\in\phi_\delta^{-1}(z), x'\in\phi_\delta^{-1}(z')} A_X(x,x') = A_Z(\phi_\delta(x),\phi_\delta(x'))$ if they are mapped to different equivalent classes. 

Consider now an arbitrary method $\ccalH$ satisfying axioms (A1)-(A2) and denote by $(Z,u_Z) = \ccalH(N_Z)$ the outcome of $\ccalH$ when applied to $N_Z$. To apply Property (P1) to this outcome we determine the minimum loop cost of $N_Z$ in the following claim.

\begin{figure}
\centering
\def \thisplotscale {0.7}
\def \unit {\thisplotscale cm}

{\scriptsize
\begin{tikzpicture}[-stealth, shorten >=2,  shorten <=2, x = \unit, y=0.9*\unit]

    \node [blue vertex, minimum size = 3*\unit] at (0,0.1) (z) {};
    \path (z) ++ (-1.3,1.3) node {\blue{\small $z$}};
    \node [point] at ( 0.2,-0.1) (1) {}; \node [point] at ( 0.7, 0.9) (2) {};
    \node [point] at (-0.6,-0.6) (3) {}; \node [point] at ( 0.6,-0.8) (4) {};
    \node [point] at (-0.8, 0.1) (5) {}; \node [point] at ( 0.9,-0.3) (6) {};
    \node [point] at (-0.6, 0.8) (7) {}; \node [point] at ( 0.1, 0.8) (8) {};
    \node [point] at ( 0.1,-0.9) (9) {}; \node [point] at ( 0.5, 0.4) (10) {}; 
    \node [point] at (-0.2, 0.3) (11) {}; \node [point] at (-0.3,-0.2) (12) {};
    \path [stealth-stealth] (3) edge (5)   (3) edge (12)  (3) edge (9)   (5) edge (11);
    \path [stealth-stealth] (1) edge (12)  (1) edge (11)  (1) edge (10)  (1) edge (6);
    \path [stealth-stealth] (4) edge (9)   (6) edge (4)   (2) edge (10)  (7) edge (11);
    \path [stealth-stealth] (8) edge (11)  (8) edge (10)  (8) edge (2)   (5) edge (7);

    \path (z) ++ (5.5,1.0) node [red vertex, minimum size = 2*\unit] (zp) {}; 
    \path (zp) ++ (1,1) node {\red{\small $z'$}};    
    \path (zp) ++ (-0.6,0.4) node [point] (1p) {};  \path (zp) ++ (0.3,-0.3) node [point] (2p) {};
    \path (zp) ++ (0.5,0.5) node [point] (3p) {};  \path (zp) ++ (-0.5,-0.5) node [point] (4p) {};
    \path [stealth-stealth] (1p) edge (3p) (3p) edge (2p) (2p) edge (4p) (4p) edge (1p);

    \path (z) ++ (3.3,-1.8) node [green vertex, minimum size = 2.1*\unit] (zpp) {}; 
    \path (zpp) ++ (1,-1) node {\green{\small $z''$}};    
    \path (zpp) ++ (-0.5,0.6) node [point] (1pp) {};  \path (zpp) ++ (0.5,-0.4) node [point] (2pp) {};
    \path (zpp) ++ (0.5,0.5) node [point] (3pp) {};  \path (zpp) ++ (-0.2,-0.7) node [point] (4pp) {};
    \path (zpp) ++ (0.1,0.1) node [point] (5pp) {};  \path (zpp) ++ (-0.5,-0.1) node [point] (6pp) {};
    \path [stealth-stealth] (5pp) edge (3pp) (5pp) edge (1pp) (5pp) edge (2pp) (5pp) edge (6pp);
    \path [stealth-stealth] (1pp) edge (6pp) (6pp) edge (4pp) (4pp) edge (2pp) (2pp) edge (3pp);    
     
    \path (2) edge [bend left, above] node {$A_Z(z,z')$} (1p);
    \path (4p) edge [bend left, above] node {$A_Z(z',z)$} (10);    
    \path (4p) edge [bend right, pos=0.52, right] node {$A_Z(z',z'')$} (3pp);    
    \path (3pp) edge [bend right, pos=0.35, right] node {$A_Z(z'',z')$} (2p);   
    \path (4) edge [bend left, below] node {$A_Z(z,z'')$} (1pp);             
    \path (6pp) edge [bend left, below] node {$A_Z(z'',z)$} (9);             
\end{tikzpicture}
} 
\vspace{-0.075in}
\caption{\small Network of equivalence classes for a given resolution. The Axiom of Transformation permits relating the clustering in the original network and the clustering in the network of equivalence classes.}
\label{fig_proof_theo_extremal_ultrametrics}
\end{figure}
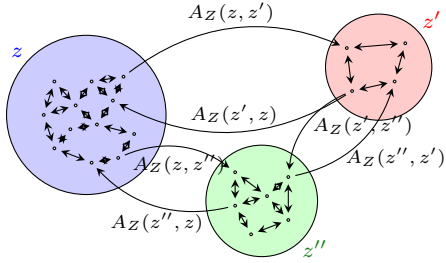

%
\begin{myclaim}\label{lemma_mlc_greater_delta}
The minimum loop cost of the network $N_Z$ is $\mlc(N_Z) > \delta$.
\end{myclaim}
\begin{myproofnoname} Assume that Claim~\ref{lemma_mlc_greater_delta} is not true, denote by $C(z, z) = [z, z', \ldots, z^{(l)}, z]$ a loop of cost smaller than $\delta$ and consider arbitrary nodes $x\in\phi_\delta^{-1}(z)$ and $x'\in\phi_\delta^{-1}(z')$. By definition, given two nodes in the same equivalence class, we can always find a chain from one to the other of cost not larger than $\delta$. Moreover, since we are assuming that $A_Z(z,z') \leq \delta$, this implies that there exists at least one node $x_1$ belonging to class $z$ and another node $x_2$ belonging to $z'$ such that $A_X(x_1, x_2) \leq \delta$. Combining these two facts, we can guarantee the existence of a chain from $x$ to $x'$ of cost not larger than $\delta$, since we can go first from $x$ to $x_1$ then from $x_1$ to $x_2$ and finally from $x_2$ to $x'$ without encountering dissimilarities greater than $\delta$. In a similar way, we can go from $x'$ to $x$ by constructing a chain that goes through all the equivalence classes in $C(z, z)$, i.e., from $z'$ to $z''$ then to $z^{(3)}$ and so on until we reach $z$. Since we can go from $x$ to $x'$ and back with chains of cost not exceeding $\delta$, it follows that $u^{\NR}_X(x,x')\leq\delta$ contradicting the assumption that $x$ and $x'$ belong to different equivalent classes. Therefore, the assumption that Claim~\ref{lemma_mlc_greater_delta} is false cannot hold. \end{myproofnoname}

%
Continuing with the main proof, since the minimum loop cost of $N_Z$ satisfies Claim~\ref{lemma_mlc_greater_delta} it follows from Property (P1) that $ u_Z(z,z') > \delta$ for all pairs of distinct equivalent classes $z,z'$. Further note that, since $\phi_\delta$ is dissimilarity reducing, Axiom (A2) implies that $u_X(x,x') \geq u_Z(z,z')$. Combining these facts, we can conclude that when $x$ and $x'$ map to different equivalence classes it holds that $u_X(x,x') \geq u_Z(z,z') >\delta$. Recall that $x$ and $x'$ mapping to different equivalence classes is equivalent to $ u^{\NR}_X(x,x')>\delta$. Consequently, we can claim that $u^{\NR}_X(x,x')>\delta$ implies $u_X(x,x')>\delta$, or, in set notation that $\{(x,x') : u^{\NR}_X(x,x')>\delta\} \subseteq \{(x,x') : u_X(x,x')>\delta\}$. Since the previous expression is true for arbitrary $\delta>0$ it implies that $u^{\NR}_X(x,x') \leq  u_X(x,x')$ for all $x, x' \in X$ as in the first inequality in \eqref{eqn_theo_extremal_ultrametrics}. \end{myproof}


\begin{myproof}[of ${\bf u_X(x,x') \leq  u^{\R}_X(x,x')}$] To prove the second inequality in \eqref{eqn_theo_extremal_ultrametrics} consider points $x$ and $x'$ with reciprocal ultrametric $u^{\R}_X(x,x') = \delta$. Let $C^*(x,x')=[x=x_0,\ldots, x_l=x']$ be a chain achieving the minimum in \eqref{eqn_reciprocal_clustering} so that we can write
\begin{equation}\label{eqn_theo_extremal_ultrametrics_pf_100}
   \delta =  u^{\R}_X(x,x') = \max_{i | x_i \in C^*(x,x')}  \,  \max \Big(A_X(x_i,x_{i+1}),  A_X(x_{i+1},x_i)\Big) .
\end{equation}
Turn attention to the symmetric two-node network $\vec{\Delta}_2(\delta,\delta)= (\{p,q\}, A_{p,q})$ with $A_{p,q}(p,q)=A_{p,q}(q,p)=\delta$ and define $(\{p,q\}, u_{p,q}) := \ccalH(\vec{\Delta}_2(\delta,\delta))$. Notice that according to Axiom (A1) we have $u_{p,q}(p,q) = \max(\delta, \delta)=\delta$.

Focus now on transformations $\phi_i:\{p,q\}\to X$ given by $\phi_i(p)=x_i$, $\phi_i(q)=x_{i+1}$ so as to map $p$ and $q$ to subsequent points in the chain $C^*(x,x')$ used in \eqref{eqn_theo_extremal_ultrametrics_pf_100}. Since it follows from \eqref{eqn_theo_extremal_ultrametrics_pf_100} that $A_X(x_i,x_{i+1})\leq\delta$  and $A_X(x_{i+1},x_i) \leq \delta$ for all $i$, it is just a simple matter of notation to observe that
\begin{align}\label{eqn_theo_extremal_ultrametrics_pf_110}
   A_X(\phi_i(p),\phi_i(q)) \! \leq \! A_{p,q}(p,q) = \delta, \,\,\, A_X(\phi_i(q),\phi_i(p)) \! \leq \! A_{p,q}(q,p) = \delta.
\end{align}
Since according to \eqref{eqn_theo_extremal_ultrametrics_pf_110} transformations $\phi_i$ are dissimilarity reducing, it follows from Axiom (A2) that $u_X(x_i,x_{i+1}) = u_X(\phi_i(p),\phi_i(q)) \leq u_{p,q}(p,q) = \delta$, for all $i$. To complete the proof we use the fact that since $u_X$ is an ultrametric and $C^*(x,x')=[x=x_0,\ldots, x_l=x']$ is a chain joining $x$ and $x'$ the strong triangle inequality dictates [cf.~\eqref{eqn_strong_triangle_inequality}] that $u_X(x,x') \leq \max_i u_X(x_i,x_{i+1}) \leq \delta$. The proof of the second inequality in \eqref{eqn_theo_extremal_ultrametrics} follows by substituting $\delta =  u^{\R}_X(x,x')$ [cf. \eqref{eqn_theo_extremal_ultrametrics_pf_100}]. \end{myproof}

\noindent Having showed both inequalities in \eqref{eqn_theo_extremal_ultrametrics}, the global proof concludes. \end{myproof}

\begin{myproof}[of Theorem \ref{theo_alternative_influence}]
Suppose there exists a clustering method $\ccalH$ that satisfies axioms (A1'') and (A2) but does not satisfy Property (P1'). This means that there exists a network $N=(X, A_X)$ with output ultrametrics $(X, u_X)=\ccalH(N)$ for which $u_X(x_1, x_2) < \sep(X, A_X)$ for at least one pair of nodes $x_1 \neq x_2 \in X$. Focus on a symmetric two-node network $\vec{\Delta}_2(s,s)=(\{p,q\}, A_{p,q})$ with $A_{p,q}(p,q)=A_{p,q}(q,p)=s = \sep(X, A_X)$ and define $(X, u_{p,q})=\ccalH(\vec{\Delta}_2(s,s))$. From Axiom (A1''), we must have that
\begin{equation}\label{eqn_alternative_axioms_imply_influence_2}
u_{p,q}(p,q)=\min \Big(\sep(X, A_X),\sep(X, A_X)\Big)=\sep(X, A_X).
\end{equation}
Construct the map $\phi:X \to \{p,q\}$ from the network $N$ to $\vec{\Delta}_2(s,s)$ that takes node $x_1$ to $\phi(x_1)=p$ and every other node $x \neq x_1$ to $\phi(x)=q$. No dissimilarity can be increased when applying $\phi$ since every dissimilarity is mapped either to zero or to $\sep(X, A_X)$ which is by definition the minimum dissimilarity in the original network [cf.~\eqref{eqn_def_separation_network}]. Hence, $\phi$ is dissimilarity reducing and from Axiom (A2) it follows that $u_X(x_1, x_2) \geq u_{p,q}(\phi(x_1), \phi(x_2)) = u_{p,q}(p,q)$. By substituting \eqref{eqn_alternative_axioms_imply_influence_2} into the previous expression, we contradict $u_X(x_1, x_2) < \sep(X, A_X)$ proving that such method $\ccalH$ cannot exist.
\end{myproof}

\begin{myproof}[of Proposition \ref{prop_unilateral_axioms}]
To show fulfillment of (A1''), consider the network $\vec{\Delta}_2(\alpha, \beta)$ and define $(\{p,q\}, u^{\U}_{p, q}) :=\ccalH^\U(\vec{\Delta}_2(\alpha, \beta))$. Since every chain connecting $p$ and $q$ must contain these two nodes as consecutive nodes, applying the definition in \eqref{eqn_unilateral_clustering_2} yields $u^{\U}_{p, q}(p,q) = \min \big(A_{p,q}(p,q), A_{p,q}(q,p)\big) = \min(\alpha,\beta)$, and Axiom (A1'') is thereby satisfied. 
In order to show fulfillment of Axiom (A2), the proof is analogous to the one developed in Proposition \ref{prop_reciprocal_axioms}. The proof only differs in the appearance of minimizations instead of maximizations to account for the difference in the definitions of unilateral and reciprocal ultrametrics [cf.  \eqref{eqn_unilateral_clustering_2} and \eqref{eqn_reciprocal_clustering}].
\end{myproof}

\begin{myproof}[of Theorem \ref{theo_unilateral_unicity}]
Given an arbitrary network $(X, A_X)$, denote by $\ccalH$ a clustering method that fulfills axioms (A1'') and (A2) and define $(X, u_X) := \ccalH(X, A_X)$. Then, we show the theorem by proving the following inequalities for all nodes $x, x' \in X$,

\begin{equation}\label{eqn_theo_unilateral_unicity_pf_010}
    u^\U_X(x,x')\leq u_X(x,x')\leq u^\U_X(x,x').
\end{equation} 

\begin{myproof}[of leftmost inequality in \eqref{eqn_theo_unilateral_unicity_pf_010}] Consider the unilateral clustering equivalence relation $\sim_{\U_X(\delta)}$ at resolution $\delta$ according to which $x \sim_{\U_X(\delta)} x'$ if and only if $x$ and $x'$ belong to the same unilateral cluster at resolution $\delta$. That is, $x \sim_{\U_X(\delta)} x'  \iff  u^{\U}_X(x,x')\leq\delta$. Further, as in the proof of Theorem \ref{theo_extremal_ultrametrics}, consider the set $Z$ of equivalence classes at resolution $\delta$. That is, $Z := X \mod \sim_{\U_X(\delta)}$. Also, consider the map $\phi_{\delta}:X\to Z$ that maps each point of $X$ to its equivalence class. Notice that $x$ and $x'$ are mapped to the same point $z$ if and only if they belong to the same block at resolution $\delta$, consequently $\phi_\delta(x) = \phi_\delta(x') \iff  u^{\U}_X(x,x')\leq\delta$. We define the network $N_Z=(Z,A_Z)$ by endowing $Z$ with the dissimilarity function $A_Z$ derived from $A_X$ as explained in \eqref{eqn_theo_extremal_ultrametrics_pf_020}. For further details on this construction, review the corresponding proof in Theorem \ref{theo_extremal_ultrametrics} and see Fig. \ref{fig_proof_theo_extremal_ultrametrics}. We stress the fact that the map $\phi_\delta$ is dissimilarity reducing for all $\delta$. 

\begin{myclaim}\label{C:sep_equiv}
The separation of the equivalence class network $N_Z$ is $\sep(N_Z) > \delta$.
\end{myclaim}
\begin{myproof} First, observe that by definition of unilateral clustering
\eqref{eqn_unilateral_clustering_2}, we know that,
\begin{equation}\label{eqn_theo_unilateral_unicity_pf_030}
    u^U_X(x,x') \leq \min(A_X(x,x'), A_X(x',x)),
\end{equation}
since a two-node chain between nodes $x$ and $x'$ is a particular chain joining the two nodes whereas the ultrametric is calculated as the minimum over all chains. Now, assume that $\sep(N_Z) \leq \delta$. Therefore, by \eqref{eqn_theo_extremal_ultrametrics_pf_020} there exists a pair of nodes $x$ and $x'$ that belong to different equivalence classes and have $A_X(x,x')\leq \delta$. However, if $x$ and $x'$ belong to different equivalence classes, they cannot be clustered at resolution $\delta$, hence, $u^U_X(x,x')>\delta$. Inequalities $A_X(x,x')\leq \delta$ and $u^U_X(x,x')>\delta$ cannot hold simultaneously since they contradict \eqref{eqn_theo_unilateral_unicity_pf_030}. Thus, it must be that $\sep(N_Z) > \delta$.\end{myproof}

Define $(Z,u_Z) := \ccalH(Z,A_Z)$ and, since $\sep(N_Z)>\delta$ (cf.~Claim~\ref{C:sep_equiv}), it follows from Property (P1') that for all $z \neq z'$ it holds $u_Z(z,z') > \delta$. Further, recalling that $\phi_\delta$ is a dissimilarity-reducing map, from Axiom (A2) we must have $u_X(x,x') \geq u_Z(\phi_\delta(x), \phi_\delta(x')) = u_Z(z, z')$ for some $z, z' \in Z$. This fact, combined with $u_Z(z,z') > \delta$, entails that when $\phi_\delta(x)$ and $\phi_\delta(x')$ belong to different equivalence classes $u_X(x,x') \geq u_Z(\phi(x),\phi(x')) >\delta$. Notice now that $\phi_\delta(x)$ and $\phi_\delta(x')$ belonging to different equivalence classes is equivalent to $ u^{\U}_X(x,x')>\delta$. Hence, we can state that $u^{\U}_X(x,x')>\delta$ implies $u_X(x,x')>\delta$ for any arbitrary $\delta>0$. In set notation, $\{(x,x') : u^{\U}_X(x,x')>\delta\} \subseteq \{(x,x') : u_X(x,x')>\delta\}$. Since the previous expression is true for arbitrary $\delta>0$, this implies that $u^{\U}_X(x,x') \leq  u_X(x,x')$, proving the left inequality in \eqref{eqn_theo_unilateral_unicity_pf_010}. \end{myproof}


\begin{myproof}[of rightmost inequality in \eqref{eqn_theo_unilateral_unicity_pf_010}] Consider two nodes $x$ and $x'$ with unilateral ultrametric value $u^{\U}_X(x,x') = \delta$. Let $C^*(x,x')=[x=x_0,\ldots, x_l=x']$ be a minimizing chain in the definition \eqref{eqn_unilateral_clustering_2} so that we can write
\begin{align}\label{eqn_theo_unilateral_unicity_pf_100}
   \delta =  u^{\U}_X(x,x') = \max_{i | x_i \in C^*(x,x')} \,  \min \Big(A_X(x_i,x_{i+1}),  A_X(x_{i+1},x_i)\Big).
\end{align}
Consider the two-node network $\vec{\Delta}_2(\delta, M)=(\{p,q\}, A_{p,q})$ where $M := \max_{x,x'} \\ A_X(x,x') $ and define $(\{p, q\}, u_{p,q}) :=\ccalH(\{p,q\},A_{p,q})$. Notice that according to Axiom (A1'') we have $u_{p,q}(p,q) = u_{p,q}(q,p) = \min( \delta, M) = \delta$, where the last equality is enforced by the definition of $M$.

Focus now on each link of the minimizing chain in \eqref{eqn_theo_unilateral_unicity_pf_100}. For every successive pair of nodes $x_i$ and $x_{i+1}$, we must have
\begin{align}
  \max \Big(A_X(x_i,x_{i+1}),  A_X(x_{i+1},x_i)\Big) \leq M, \label{eqn_theo_unilateral_unicity_pf_101} \\
  \min \Big(A_X(x_i,x_{i+1}),  A_X(x_{i+1},x_i)\Big) \leq \delta.  \label{eqn_theo_unilateral_unicity_pf_102}
\end{align}
Expression \eqref{eqn_theo_unilateral_unicity_pf_101} is true since $M$ is defined as the maximum dissimilarity in $A_X$. Inequality \eqref{eqn_theo_unilateral_unicity_pf_102}
is justified by \eqref{eqn_theo_unilateral_unicity_pf_100}, since $\delta$ is defined as the maximum among links of the minimum distance in both directions of the link.
This observation allows the construction of dissimilarity-reducing maps $\phi_i:\{p,q\}\to X$,

\begin{equation}\label{eqn_theo_unilateral_unicity_pf_103}
\phi_i :=
\begin{cases}
\phi_i(p)=x_i, \phi_i(q)=x_{i+1}, \quad \text{if} \,\, \hat{A}_X(x_i,x_{i+1}) = A_X(x_i,x_{i+1}) \\
\phi_i(q)=x_i, \phi_i(p)=x_{i+1},  \quad \text{otherwise.}
\end{cases}
\end{equation}

In this way, we can map $p$ and $q$ to subsequent nodes in the chain $C(x,x')$ used in \eqref{eqn_theo_unilateral_unicity_pf_100}. Inequalities \eqref{eqn_theo_unilateral_unicity_pf_101} and \eqref{eqn_theo_unilateral_unicity_pf_102} combined with the map definition in \eqref{eqn_theo_unilateral_unicity_pf_103} guarantee that $\phi_i$ is a dissimilarity-reducing map for every $i$. 
Since clustering method $\ccalH$ satisfies Axiom (A2), it follows that 
\begin{align}\label{eqn_theo_unilateral_unicity_pf_120}
   u_X(\phi_i(p),\phi_i(q)) \leq u_{p,q}(p,q) = \delta, \quad \forall\ i.
\end{align}
Substituting $\phi_i(p)$ and $\phi_i(q)$ in \eqref{eqn_theo_unilateral_unicity_pf_120} by the corresponding nodes given by the definition \eqref{eqn_theo_unilateral_unicity_pf_103}, we can write $u_X(x_i, x_{i+1})= u_X(x_{i+1}, x_i) \leq \delta$, for all $i$, where the symmetry property of ultrametrics was used.
To complete the proof we invoke the strong triangle inequality \eqref{eqn_strong_triangle_inequality} and apply it to $C(x,x')=[x=x_0,\ldots, x_l=x']$, the minimizing chain in \eqref{eqn_theo_unilateral_unicity_pf_100}. As a consequence, $u_X(x,x') \leq \max_i u_X(x_i,x_{i+1}) \leq \delta$. The proof of the right inequality in \eqref{eqn_theo_unilateral_unicity_pf_010} is completed by substituting $\delta =  u^{\U}_X(x,x')$ [cf. \eqref{eqn_theo_unilateral_unicity_pf_100}] into the last previous expression. \end{myproof}

\noindent Having proved both inequalities in \eqref{eqn_theo_unilateral_unicity_pf_010}, unilateral clustering is the only method that satisfies axioms (A1'') and (A2), completing the global proof.
\end{myproof}

\begin{myproof}[of Theorem \ref{theo_extremal_ultrametrics_2}]
The leftmost inequality in \eqref{eqn_theo_extremal_ultrametrics_2} can be proved using the same method of proof used for the leftmost inequality in \eqref{eqn_theo_unilateral_unicity_pf_010} within the proof of Theorem \ref{theo_unilateral_unicity}. The proof of the rightmost inequality in \eqref{eqn_theo_extremal_ultrametrics_2} is equivalent to the proof of the rightmost inequality in Theorem \ref{theo_extremal_ultrametrics}.
\end{myproof}

{\small

\bibliographystyle{spbasic}      
\bibliography{clustering_biblio.bib}   
}

\end{document}